%% file: main.tex
\documentclass[10pt,twocolumn,letterpaper]{article}

\usepackage[pagenumbers]{cvpr} 

\input{preamble}

\definecolor{cvprblue}{rgb}{0.21,0.49,0.74}
\usepackage[pagebackref,breaklinks,colorlinks,allcolors=cvprblue]{hyperref}
\usepackage{ulem} 

\def\paperID{3722} 
\def\confName{CVPR}
\def\confYear{2025}

\title{\ourfullname: Real-time High-fidelity Radiance Field Rendering}

\author{
Cheng Sun$^{1}$ \quad\;
Jaesung Choe$^{1}$ \quad\;
Charles Loop$^{1}$ \quad\;
Wei-Chiu Ma$^{2}$ \quad\;
Yu-Chiang Frank Wang$^{1,3}$ \\
{\small $^{1}$NVIDIA} \quad
{\small $^{2}$Cornell University} \quad
{\small $^{3}$National Taiwan University}
}

\begin{document}
\input{sec/0_teaser}
\input{sec/0_abstract}
\input{sec/1_intro}
%
\input{sec/2_related}
\input{sec/3_0_approach}
\input{sec/3_1_0_SVRaster}
\input{fig/scene_repr}
\input{sec/3_1_1_scene_repr}

\input{fig/rasterization}
\input{fig/order}
\input{sec/3_1_2_rasterization}
\input{sec/3_2_0}
\input{fig/voxel_init}
\input{sec/3_2_1_scene_init}
\input{sec/3_2_2_scene_adaptive}
\input{sec/3_2_3_optim_objective}
\input{sec/3_2_4_tsdf_mc}
\input{fig/exp_mipnerf360}
\input{fig/exp_meshes}
\input{tab/mipnerf360}
\input{tab/tant_db}
\input{fig/failuare_case}
\input{sec/4_0_experiments}

\input{sec/4_1_impl_details}
\input{tab/model_size_scaled_fps}
\input{sec/4_2_nvs}
\input{tab/main_abla}
\input{sec/4_3_abla}
\input{tab/meshes}
\input{sec/4_4_mesh}
\input{sec/4_5_feat_fusion}

\input{sec/5_conclusion}

\paragraph{Acknowledgements.}
We thank Min-Hung Chen and Yueh-Hua Wu for helpful paper proofreading.
\paragraph{Updates.}
\begin{itemize}
\item Mar 2025: (1) Revise literature review. (2) Add Scannet++ results.
\item Feb 2025: (1) Code release. (2) Add more volume fusion examples. (3) Novel-view synthesis quality improved. (4) Provide fast rendering and fast training variants. (5) Discuss results more.
\item Dec 2024: Paper release on arxiv.
\end{itemize}
{
    \small
    \bibliographystyle{ieeenat_fullname}
    \bibliography{main}
}
\clearpage
\maketitlesupplementary
\renewcommand\thesection{\Alph{section}}
\setcounter{section}{0}
\input{supp_material/main}

\end{document}

%% file: preamble.tex
\usepackage{multirow}
\usepackage{graphicx}
\usepackage{subcaption}
\usepackage{dsfont}
\usepackage{physics}
\usepackage{tikz}
\usepackage[permil]{overpic}
\usepackage{makecell}

\usepackage[frozencache,cachedir=.]{minted}
\definecolor{gold}{rgb}{1.0, 0.874, 0}
\definecolor{silver}{rgb}{0.75, 0.75, 0.75}
\definecolor{brown}{rgb}{0.8, 0.678, 0.4}
\definecolor{gray}{rgb}{0.5, 0.5, 0.5}
\newcommand{\gold}[1]{\colorbox{gold}{\textbf{#1}}}
\newcommand{\silver}[1]{\colorbox{silver}{{#1}}}
\newcommand{\brown}[1]{\colorbox{brown}{{#1}}}

\makeatletter
\DeclareRobustCommand\onedot{\futurelet\@let@token\@onedot}
\def\@onedot{\ifx\@let@token.\else.\null\fi\xspace}
\def\eg{\textit{e.g}\onedot} 
\def\ie{\textit{i.e}\onedot}

\usepackage{enumitem}
\setlist[itemize]{parsep=0pt,partopsep=0pt,leftmargin=*,itemsep=5pt}
\setlist[enumerate]{parsep=0pt,partopsep=0pt,leftmargin=*,itemsep=5pt}
\makeatletter
\renewcommand\paragraph{\@startsection{paragraph}{4}{\z@}%
    {.5em}%
    {-1em}%
    {\normalfont\normalsize\bfseries}}
\makeatother
\expandafter\def\expandafter\normalsize\expandafter{%
    \normalsize%
    \setlength\abovedisplayskip{1ex}%
    \setlength\belowdisplayskip{1ex}%
    \setlength\abovedisplayshortskip{1ex}%
    \setlength\belowdisplayshortskip{1ex}%
}

\newcommand{\ourfullname}{Sparse Voxels Rasterization\xspace}
\newcommand{\ourshortname}{SVRaster\xspace}

\newcommand{\Real}{\mathbb{R}}

\newcommand{\Realprob}{\mathbb{R}_{\in[0,1]}}

\newcommand{\maxlv}{L}
\newcommand{\nsamp}{K}
\newcommand{\nshd}{N_{\mathrm{shd}}}

\newcommand{\ncam}{N_{\mathrm{cam}}}

\newcommand{\ray}{\mathbf{r}}
\newcommand{\rayo}{\ray_{\mathrm{o}}}
\newcommand{\rayd}{\ray_{\mathrm{d}}}
\newcommand{\rayq}{\mathbf{q}}
\newcommand{\pt}{\mathbf{p}}

\newcommand{\vox}{\mathbf{v}}

\newcommand{\voxcen}{\vox_{\mathrm{c}}}
\newcommand{\voxsiz}{\vox_{\mathrm{s}}}
\newcommand{\voxgeo}{\vox_{\mathrm{geo}}}
\newcommand{\voxshs}{\vox_{\mathrm{sh}}}

\newcommand{\voxinterval}{\vox_{\mathrm{interval}}}
\newcommand{\voxrate}{\vox_{\mathrm{rate}}}
\newcommand{\voxpriority}{\vox_{\mathrm{priority}}}

\newcommand{\bigv}{\mathbf{V}}

\newcommand{\pixrgb}{\mathrm{C}}
\newcommand{\loss}{\mathcal{L}}

\newcommand{\hyper}{h}
\newcommand{\hyperss}{\hyper_{\mathrm{ss}}}
\newcommand{\hyperinitgeo}{\hyper_{\mathrm{geo}}}
\newcommand{\hyperinitlv}{\hyper_{\mathrm{lv}}}
\newcommand{\hyperoutlv}{\hyper_{\mathrm{out}}}
\newcommand{\hyperoutnumscale}{\hyper_{\mathrm{ratio}}}
\newcommand{\hyperevery}{\hyper_{\mathrm{every}}}
\newcommand{\hyperprune}{\hyper_{\mathrm{prune}}}
\newcommand{\hyperpercent}{\hyper_{\mathrm{percent}}}
\newcommand{\hypertstop}{\hyper_{\mathrm{T}}}
\newcommand{\hyperrate}{\hyper_{\mathrm{rate}}}

\newcommand{\explin}{\mathrm{explin}}
\newcommand{\softplus}{\mathrm{softplus}}

\newcommand{\interp}{\mathrm{interp}}
\newcommand{\sheval}{\mathrm{sh\_eval}}
\newcommand{\normalize}{\mathrm{normalize}}
\newcommand{\seglen}{l}

\newcommand{\world}{\mathbf{w}}
\newcommand{\worldcen}{\world_{\mathrm{c}}}
\newcommand{\worldsiz}{\world_{\mathrm{s}}}

%% file: sec/0_teaser.tex
\twocolumn[{%
    \maketitle
    \renewcommand\twocolumn[1][]{#1}%
    \vspace{-7mm} 
    \centering
    \includegraphics[width=.99\linewidth]{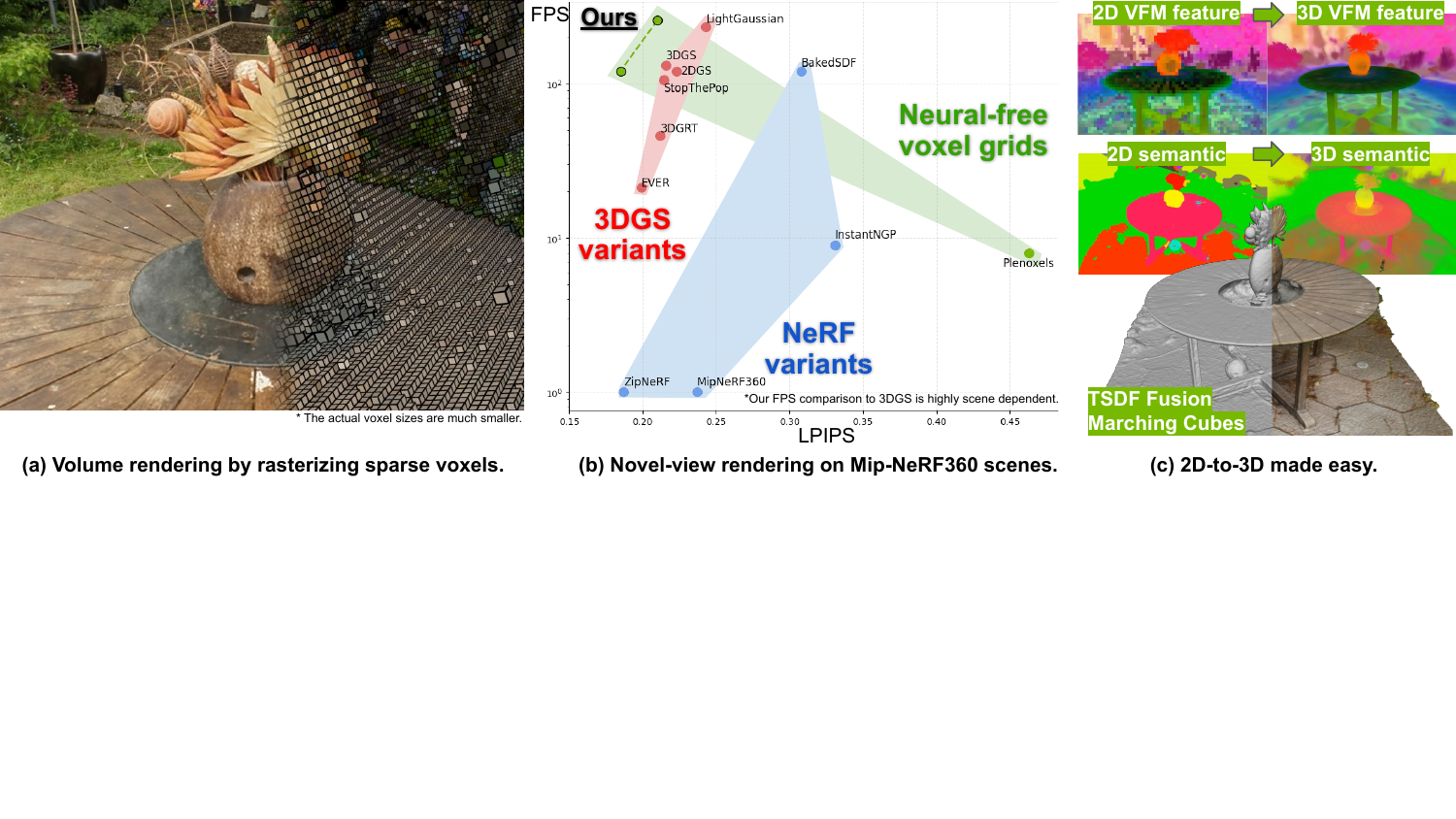}
    \vspace{-0.5em}
    \captionof{figure}{
    We propose \textbf{\ourshortname}, a novel framework for multi-view reconstruction and novel view synthesis. \textbf{(a)} Sparse voxel representation effectively captures the volume density and radiance field of the scene, without the need for neural networks, 3D Gaussians, and sparse-points prior. \textbf{(b)} Using our customized sparse voxel rasterizer, we can learn the underlying 3D scene efficiently and achieve state-of-the-art performance in both rendering quality and speed. \textbf{(c)} Notably, lifting 2D modal to the trained sparse voxels is simple and efficient by integrating the classic Volume Fusion~\cite{volumefusion,kinectfusion,bundlefusion}. We show examples of vision foundation model feature field from RADIO~\cite{radio}, semantic field from Segformer~\cite{segformer}, and signed distance field from rendered depth, making it flexible and suitable for a wide range of applications.
    }
    \label{fig:teaser}    
    \vspace{7mm}
}]

%% file: sec/0_abstract.tex
\begin{abstract}
We propose an efficient radiance field rendering algorithm that incorporates a rasterization process on adaptive sparse voxels without neural networks or 3D Gaussians. 
There are two key contributions coupled with the proposed system.
The first is to adaptively and explicitly allocate sparse voxels to different levels of detail within scenes, faithfully reproducing scene details with $65536^3$ grid resolution while achieving high rendering frame rates.
Second, we customize a rasterizer for efficient adaptive sparse voxels rendering. We render voxels in the correct depth order by using ray direction-dependent Morton ordering, which avoids the well-known popping artifact found in Gaussian splatting. 
Our method improves the previous neural-free voxel model by over 4db PSNR and more than 10x FPS speedup, achieving state-of-the-art comparable novel-view synthesis results.
Additionally, our voxel representation is seamlessly compatible with grid-based 3D processing techniques such as Volume Fusion, Voxel Pooling, and Marching Cubes, enabling a wide range of future extensions and applications.
Code: \href{https://github.com/NVlabs/svraster}{github.com/NVlabs/svraster}
\end{abstract}

%% file: sec/1_intro.tex
\section{Introduction}
\label{sec:intro}

Gaussian splatting~\cite{3dgs} has emerged as one of the most promising solutions for novel view synthesis. It has drawn wide attention across multiple communities due to its exceptional rendering speed and ability to capture the nuanced details of a scene. Nevertheless, Gaussian splatting in its base form has two key limitations: first, sorting Gaussians based on their centers does not guarantee proper depth ordering. It may result in popping artifacts~\cite{stopthepop} 
(\ie, sudden color changes for consistent geometry) when changing views. Second, the volume density of a 3D point is ill-defined when covered by multiple Gaussians. This ambiguity makes surface reconstruction non-trivial.

On the other hand, grid representations inherently avoid these issues---both ordering and volume are well-defined. However, due to ray casting, the rendering speed of these methods~\cite{dvgo,tensorf,eg3d,neuralpbir,voxurf,neuralangelo} is 
slower than that of Gaussian Splatting. This raises the question: is it possible to take the best of both worlds? Can we combine the efficiency of Gaussian splatting with the well-defined grid properties?

Fortunately, the answer is yes. Our key insight is to revisit voxels---a well-established primitive with decades of history. Voxel representations are inherently compatible with modern graphics engines and can be rasterized efficiently. Additionally, previous work has demonstrated their ability to model scene volume densities, despite through volume ray casting. This positions voxels as the perfect bridge between rasterization and volumetric representation. 
However, naively adopting voxels does not work well~\cite{plenoxels}. Since a scene may consist of different levels of detail.

With these observations in mind, we present \ourshortname, a novel framework that combines the efficiency of rasterization in 3DGS with the structured volumetric approach of grid-based representations.
\ourshortname leverages (1) multi-level sparse voxels to model 3D scenes and (2) implements a direction-dependent Morton order encoding that facilitates the rasterization rendering from our adaptive-sized sparse voxel representation.
Our rendering is free from popping artifacts because the 3D space is partitioned into disjoint voxels, and our sorting ensures the correct rendering order.
Moreover, thanks to the volumetric nature and neural-free representation of \ourshortname, our sparse voxel grid can be easily and seamlessly integrated with classical grid-based 3D processing algorithms.

We show that our method is training fast, rendering fast, and achieves novel-view synthesis quality comparable to the state-of-the-art.
We also integrate Volume Fusion, Voxel Pooling, and Marching Cubes operations into our adaptive sparse voxels, which showcases promising results of mesh extraction and 2D foundation feature fusion.

%% file: sec/2_related.tex
\section{Related Work}
\label{sec:preliminary}

Differentiable volume rendering has made significant strides in 3D scene reconstruction and novel-view synthesis tasks. Neural Radiance Fields (NeRF)~\cite{nerf} laid the foundation for this progress by optimizing a volumetric function, parameterized by multi-layer perceptrons (MLPs) to encode both geometry and appearance through differential rendering.
Subsequent studies focus on accelerating speed by decomposing large MLPs into grid-based representations, with a shallow MLP typically still being employed.
Several grid representations have been explored---dense grids~\cite{dvgo}, factorized grids~\cite{tensorf}, tri-planes~\cite{eg3d}, and hash grids~\cite{instant-ngp}.

In particular, our method is closely related to sparse voxel grid representations.
Sparse Voxel Octrees~\cite{svoctree} and VDB trees~\cite{vdb} are commonly used to manage sparse voxels and facilitate rendering.
In contrast, our sparse voxels are stored in a 1D array without advanced data structures.
Our rasterizer with the proposed direction-dependent Morton order ensures correct rendering order.
To model volumes in sparse leaf nodes, different strategies exist, such as low-resolution 3D grid~\cite{vdb,fVDB} or implicit neural field~\cite{nsvf,nglod}.
Our leaf node is a single voxel with explicit density and color parameters, like Plenoxels~\cite{plenoxels}.
Regarding voxel levels, previous methods allocate voxels to a target level~\cite{nsvf,nglod,plenoctree,plenoxels} or use a shallow tree~\cite{vdb,fVDB}.
Our voxels can adaptively fit into different levels across the entire tree depth, maximizing flexibility and scalability.

Another scalability issue of the previous grid-based methods is that they still use some sort of dense 3D grid in the field of novel-view synthesis.
For instance, dense occupancy grids~\cite{dvgo,tensorf,instant-ngp,nerfacc} for free-space skipping or dense pointer grid~\cite{plenoxels} to support sparse voxel lookup.
We do not use any 3D dense data structures.

3D Gaussian Splatting (3DGS)~\cite{3dgs} takes a different approach by representing scenes with 3D Gaussian primitives and using rasterization for rendering, achieving state-of-the-art trade-off for quality and speed. Our work is inspired by the efficiency of using a rasterizer by 3DGS. However, 3DGS exhibits view-inconsistent popping artifacts due to inaccurate rendering order and primitive overlapping. Some recent works mitigate this artifact~\cite{stopthepop,3dgrt}, but completely resolving it significantly harms rendering speed~\cite{ever}.
Our method does not suffer from the popping artifact.
Typically, 3DGS~\cite{3dgs} initializes Gaussians from the triangulated points via COLMAP~\cite{colmap}, which is especially critical for unbounded scenes. The sparse points geometry is also employed to guide the training of volumetric representations and has shown improvement in some NeRF-based methods~\cite{ds-nerf,geoneus}.
In this work, we do not use the sparse points.

Mesh reconstruction is an important extended topic for both volumetric-based and GS-based methods. Recent 3DGS variants~\cite{2dgs,sugar} with proper regularization have shown a good speed-accuracy trade-off. However, directly extracting meshes from 3DGS remains challenging, so these methods still rely on volumetric-based post-processing algorithms to extract meshes.
NeRF variants~\cite{neus,volsdf,neuralpbir,neuralangelo,voxurf,neus2} improve surface quality by rendering a signed distance field (SDF) instead of density field. The well-defined volume makes it straightforward to extract a mesh from the isosurface.
Our sparse voxels are also a volumetric representation. Though our voxel also follows volumetric representation focusing on the density field, we still can achieve promising accuracy with minimal time.

For efficient rendering, another line of research is to convert (or distill) the geometry and appearance of a high-quality but slow model into other efficient kind of representations such as smaller MLPs~\cite{kilonerf,hybridnerf}, meshes~\cite{adaptive-shell,baked-sdf,binarygrid}, grids~\cite{smerf}, or the recent Gaussians~\cite{radsplat}. Typically, the powerful Zip-NeRF~\cite{zipnerf} is employed as the teacher model which requires hours of training per scene.
Our volumetric representation could enable training-free conversion from a large implicit field, which is however out of our current scope.

Finally, the sparse voxel is also a widely-used representation for 3D processing~\cite{poissonrecon,volumefusion,kinectfusion,bundlefusion,voxelhashing,bayesianfusion}
and understanding~\cite{mink,spconv,fVDB}.
We mainly focus on rendering in this work while we believe that adapting our method to these techniques is a promising future direction.

%% file: sec/3_0_approach.tex
\section{Approach}

We present our approach as follows.
First in \cref{ssec:svraster}, we introduce our sparse voxels scene representation and our rasterizer for rendering sparse voxels into pixels.
Later in \cref{ssec:pg_sv_optim}, we describe the progressive scene optimization strategy, which is designed for faithfully reconstructing the scene from multi-view images using our sparse voxels.

%% file: sec/3_1_0_SVRaster.tex
\subsection{\ourfullname}
\label{ssec:svraster}

Recent neural radiance field rendering approaches, such as NeRF~\cite{nerf} and 3DGS~\cite{3dgs} variants, use the following alpha composition equation to render a pixel's color~$\pixrgb$:
\begin{align} \label{eq:alpha_blend}
    \pixrgb &= \sum\nolimits_{i=1}^{N} T_i \cdot \alpha_i \cdot c_i ~, &
    T_i &= \prod\nolimits_{j=1}^{i-1} \left(1 - \alpha_j\right) ~,
\end{align}
where $\alpha_i \in \Realprob$ and $c_i \in \Realprob^{3}$ are alpha and view-dependent color at the $i$-th sampled point or primitive on the pixel-ray. The quantity $T_i$ is known as the transmittance. 
At a high-level, the difference between each method comes down to: {\it i)} how they find the $N$ points or primitives to composite a pixel-ray and {\it ii)} how the rgb and alpha values are determined from their scene representations.

Our sparse-voxel method follows the same principle.
In \cref{sssec:scene_repr}, we provide details of our sparse voxels scene representation and how the $\alpha_i$ and $c_i$ in \cref{eq:alpha_blend} are computed from a voxel.
In \cref{sssec:rasterization}, we present our rasterizer, that gathers the $N$ voxels to be composited for each pixel.

%% file: fig/scene_repr.tex
\begin{figure}
    \centering
    \includegraphics[width=.999\linewidth]{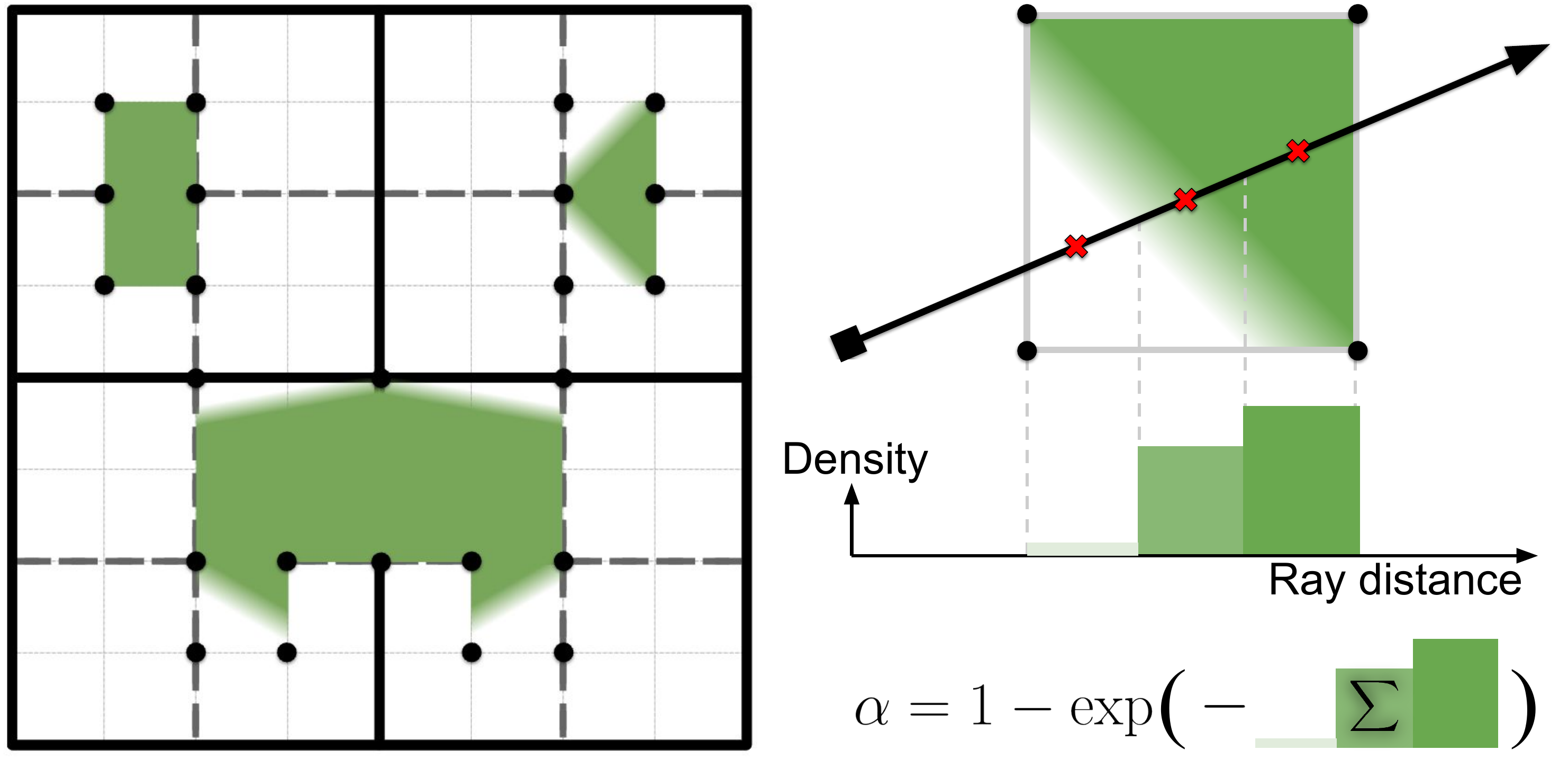}
    \begin{subfigure}{0.48\linewidth}
        \centering
        \caption{}
        \label{fig:sparse_voxels}
    \end{subfigure}
    \hfill
    \begin{subfigure}{0.48\linewidth}
        \centering
        \caption{}
        \label{fig:vox_rend}
    \end{subfigure}
    \caption{
        {\bf Sparse voxels scene representation.}
        {\bf (Left)} We allocate voxel under an Octree layout.
        Each voxels has its own Spherical Harmonic coefficient for view-dependent appearance.
        The color field is approximated as a constant inside a voxel when rendering a view for efficiency.
        The density field is trilienarly varied inside a voxel and is modeled by the density values on the corner grid points (\ie, the black dots $\bullet$) of each voxel.
        The grid points densities are shared between adjacent voxels.
        {\bf (Right)} We evenly sample $\nsamp$ points inside the segment of ray-voxel intersection to compute volume integration for its alpha value contributing to the pixel ray.
        See \cref{sssec:scene_repr} for details.
    }
    \label{fig:scene_repr}
    \vspace{-1em}
\end{figure}

%% file: sec/3_1_1_scene_repr.tex
\subsubsection{Scene Representation}
\label{sssec:scene_repr}

We first describe the grid layout for allocating our sparse voxels and then derive the alpha value, view-dependent color, and other geometric properties needed for the composite rendering of a voxel.

\paragraph{Sparse voxel grid.}
Our \ourshortname constructs 3D scenes using a sparse voxel representation.
We allocate voxels following an Octree space partition rule (\ie, \textbf{Octree layout} as illustrated in \cref{fig:sparse_voxels}) for two reasons necessary for achieving high-quality results.
First, it facilitates the correct rendering order of voxels with various sizes (\cref{sssec:rasterization}).
Second, we can adaptively fit the sparse voxels to different scene level-of-details (\cref{ssec:pg_sv_optim}).
Note that our representation does not replicate a traditional \textbf{Octree data structure} with parent-child pointers or linear Octree.
Specifically, we only keep voxels at the Octree leaf nodes without any ancestor nodes.
Our sorting-based rasterizer will project voxels to image space and guarantee all voxels are in the correct order when rendering.
In sum, we store individual voxels in arbitrary order without the need to maintain a more complex data structure, thanks to the flexibility provided by our rasterizer.


We choose a maximum level of detail $\maxlv$ 
($= 16$ in this work) that defines a maximum grid resolution at $(2^{\maxlv})^3$. Let $\worldsiz \in \Real$ be the Octree size and $\worldcen \in \Real^3$ be the Octree center in the world space.
The voxel index $v\!=\!\{i,j,k\} \in [0,\ldots,2^{\maxlv}\!\!\!-\!\!1]^3$ together with a Octree level $l\in[1,\maxlv]$ ($l=0$ represent root node and is not used) define voxel size $\voxsiz$ and voxel center $\voxcen$ as:
\begin{align} \label{eq:vox_coord}
    \voxsiz &= \worldsiz \cdot 2^{-l} ~, &
    \voxcen &= \worldcen - 0.5 \cdot \worldsiz + \voxsiz \cdot v ~.
\end{align}
Internally, we map the grid index to its Morton code using a well-known bit interleaving operation in the low-level CUDA implementation. Please see supplementary materials for more details.



\paragraph{Voxel alpha from density field.}
Next, we present details for the geometry and appearance modeling of each voxel primitive.
For scene geometry, we use eight parameters corresponding to the voxel corners to model a trilinear density field inside each voxel, denoted as $\voxgeo \in \Real^{2\times 2\times 2}$.
Sharing corners among adjacent voxels results in a continuous density field.

We also need an activation function to ensure a non-negative density value for the raw density from $\voxgeo$.
For this purpose, we use exponential-linear activation:
\begin{equation} \label{eq:explin}
    \explin(x) = \begin{cases}
        x & \text{if } x > 1.1\\
        \exp\left(\frac{x}{1.1} -1+\ln1.1\right)& \text{otherwise}
    \end{cases} ~,
\end{equation}
which approximates $\softplus$ but is more efficient to compute.
For a sharp density field inside a voxel, we apply the non-linear activation after trilinear interpolation~\cite{dvgo,relufield}.

To derive the alpha value of a voxel contributing to the alpha composition formulation in \cref{eq:alpha_blend}, we evenly sample $\nsamp$ points in the ray segment of ray-voxel intersection as depicted in \cref{fig:vox_rend}.
The equation follows the numerical integration for volume rendering as in NeRF~\cite{nerf,Max}:
\begin{equation} \label{eq:vox_alpha}
    \alpha = 1 - \exp\left( - \frac{\seglen}{\nsamp} \sum_{k=1}^{\nsamp} \explin\left(\interp\left(\voxgeo, \rayq_k\right)\right) \right) ~,
\end{equation}
where $\seglen$ is the ray segment length, $\rayq_k$ is the local voxel coordinate of the $k$-th sample point, and $\interp(\cdot)$ indicates trilinear interpolation.

\paragraph{Voxel view-dependent color from SHs.}
To model view-dependent scene appearance, we use $\nshd$ degree SH.
For increased efficiency, we assume the SH coefficients stay constant inside a voxel, denoted as $\voxshs \in \Real^{(\nshd+1)^2\times 3}$.
We approximate voxel colors as a function of the direction from the camera position $\rayo$ to the voxel center $\voxcen$ instead of individual ray direction $\rayd$ for the sake of efficiency following 3DGS:
\begin{equation} \label{eq:vox_color}
    c = \max(0, \sheval(\voxshs, \normalize(\voxcen - \rayo))) ~,
\end{equation}
which is the view-dependent color intensity of the voxel contributing to the pixel composition \cref{eq:alpha_blend}.
Due to the approximation, the resulting SH color of a voxel can be shared by all covered pixels in the image rather than evaluating the SH for each intersecting ray.

\paragraph{Voxel normal.}
The rendering of other features or properties is similar to rendering a color image by replacing the color term $c$ in \cref{eq:alpha_blend} with the target modality like the normal of a voxel density field.
For rendering efficiency, we assume the normal stays constant inside a voxel, which is represented by the analytical gradient of the density field at the voxel center:
\begin{equation} \label{eq:vox_normal}
    \vec{\bf n} = \normalize\left( \nabla_{\rayq} ~\interp\left(\voxgeo, \rayq_{\mathrm{c}}\right) \right) ~,
\end{equation}
where $\rayq_{\mathrm{c}}=(0.5,0.5,0.5)$ and the closed-form equations for forward and backward passes are in the supplementary material. 
Similar to the SH colors, the differentiable voxel normals are computed once in pre-processing and are shared by all the intersecting rays in the image.

\paragraph{Voxel depth.}
Unlike colors and normals, the point depths to composite is efficient to compute so we do the same $K$ points sampling as in the voxel alpha value in \cref{eq:vox_alpha} for a more precise depth rendering.
We manually expand and simplify the forward and backward computation for small number of $K$ in our CUDA implementation.
Please refer to supplementary materials for details.

%% file: fig/rasterization.tex
\begin{figure}
    \centering
    \includegraphics[width=\linewidth]{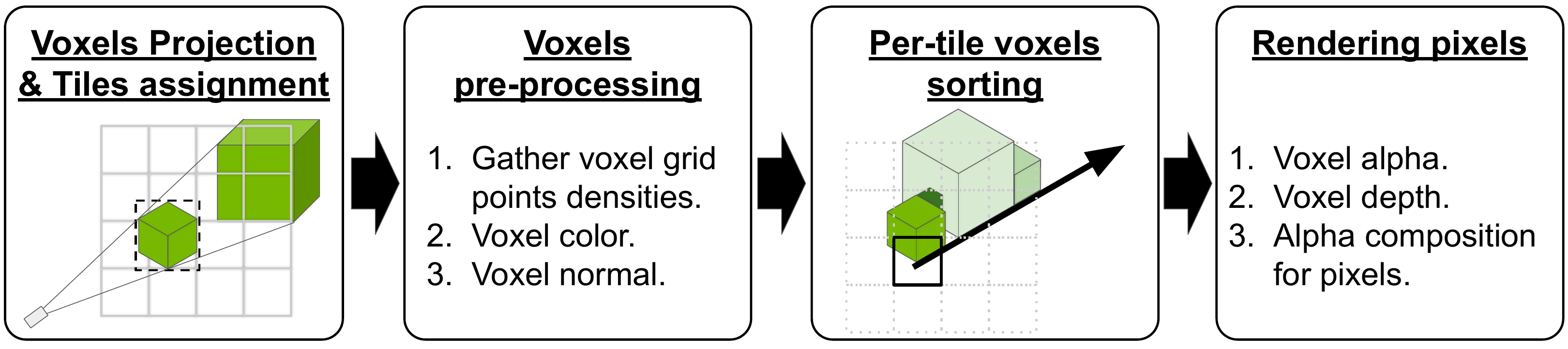}
    \caption{
        {\bf Rasterization procedure.}
        Refer to \cref{sssec:rasterization} for details.
    }
    \label{fig:rasterization}
    \vspace{-.5em}
\end{figure}

%% file: fig/order.tex
\begin{figure}
    \centering
    \begin{subfigure}{0.17\linewidth}
        \centering
        \includegraphics[width=\linewidth]{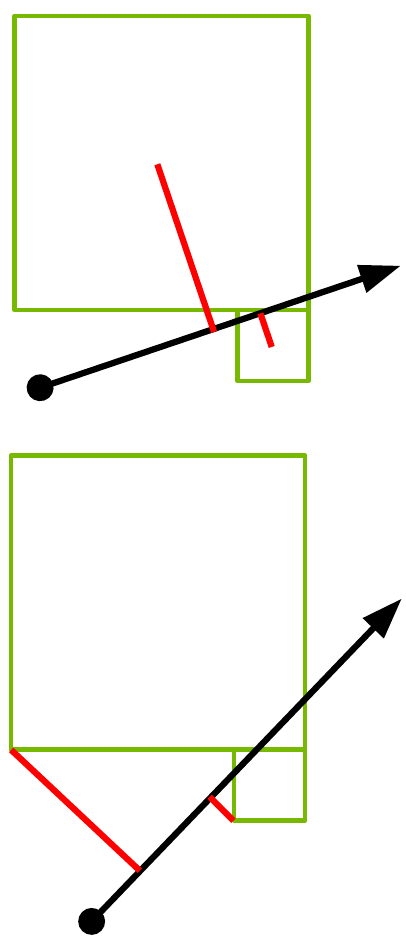}
        \caption{}
        \label{fig:incorrect_order}
    \end{subfigure}
    \hfill
    \begin{subfigure}{0.38\linewidth}
        \centering
        \includegraphics[width=\linewidth]{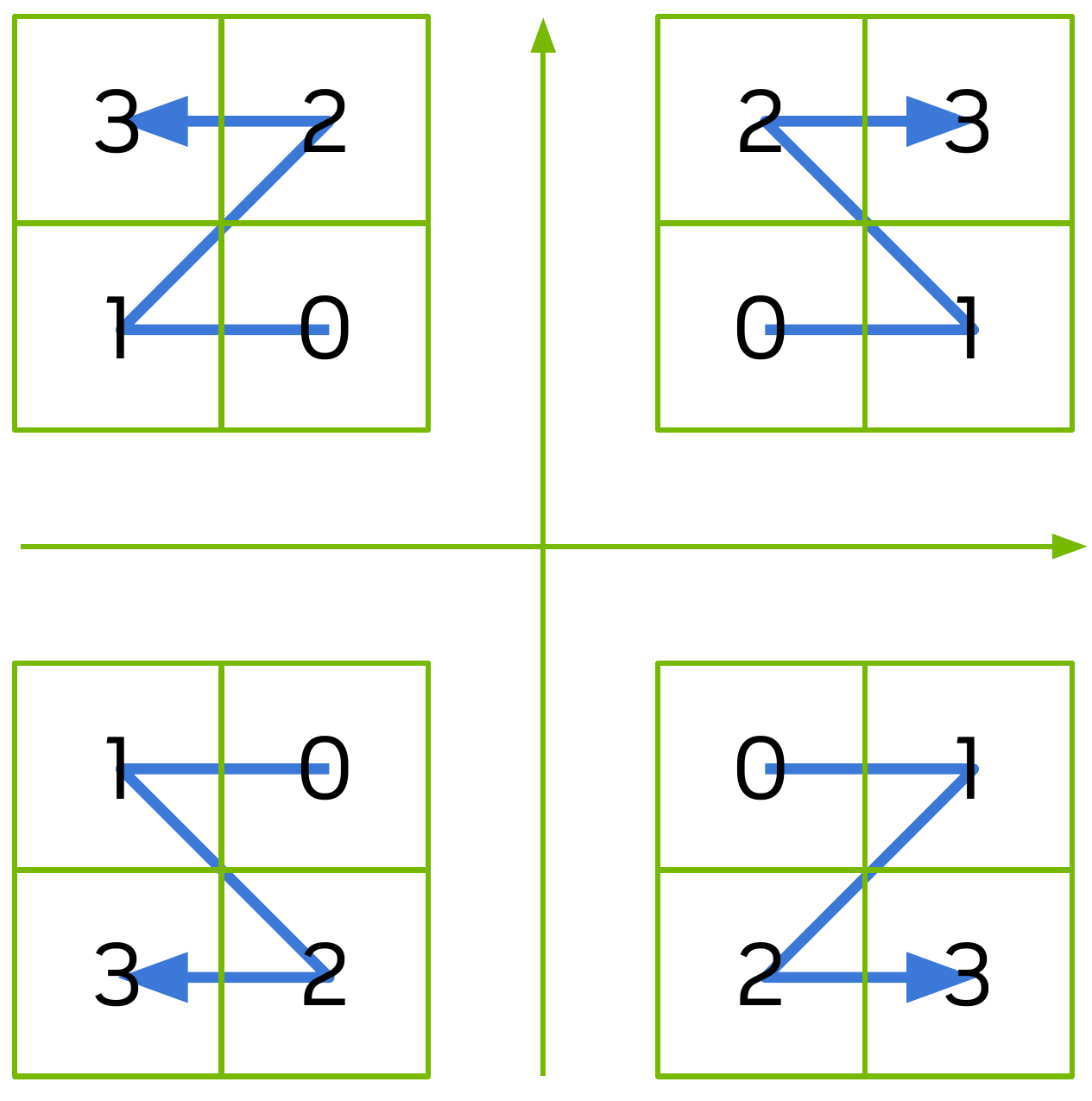}
        \caption{}
        \label{fig:cam_quad_and_order}
    \end{subfigure}
    \hfill
    \begin{subfigure}{0.38\linewidth}
        \centering
        \includegraphics[width=\linewidth]{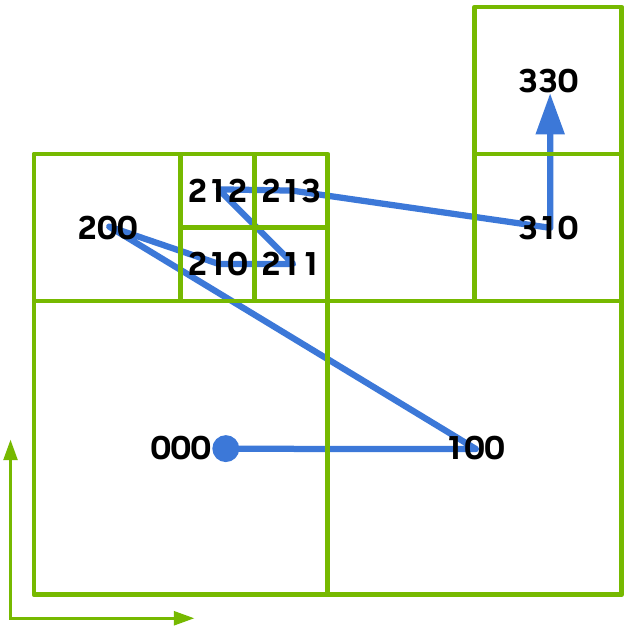}
        \caption{}
        \label{fig:order_example}
    \end{subfigure}
    \caption{
        {\bf Illustration of the rendering order.}
        {\bf (a)} In both cases, the smaller voxels should be rendered first but they will arranged behind the larger voxels if using voxel centers or the nearest corners as the sorting order.
        {\bf (b)} We show the four types of Morton order under the 2D world. The voxel rendering order under an Octree node is depend on which world quadrant the ray direction is pointing to.
        {\bf (c)} An toy example of sorting the Morton order encoding. All the ray directions going toward the up-right quadrant can use the sorted voxels for a correct rendering order.
        See \cref{sssec:rasterization} for details.
    }
    \label{fig:order}
    \vspace{-.5em}
\end{figure}

%% file: sec/3_1_2_rasterization.tex
\subsubsection{Rasterization Algorithm}
\label{sssec:rasterization}

The overview of our sparse voxel rasterization procedure is depicted in \cref{fig:rasterization}.
We build our sorting-based rasterizer based on the highly efficient CUDA implementation of 3DGS~\cite{3dgs}.
The procedure is detailed in the following.

\paragraph{Projection to image space.}
The first step of rasterization is projecting sparse voxels onto image space and assigning the voxels to the tiles (\ie, image patches) they cover.
In practice, we project the eight corner points of each voxel.
The voxel is assigned to all tiles overlapped with the axis-aligned bounding box formed by the projected eight points.

\paragraph{Pre-processing voxels.}
For active voxels assigned to tiles of the target view, we gather their densities $\voxgeo$ from the grid points, compute view-dependent colors from their spherical harmonic coefficients with \cref{eq:vox_color}, and derive voxel normals by \cref{eq:vox_normal}.
The pre-processed voxel properties are shared among all pixels during rendering.

\paragraph{Sorting voxels.}
For accurate rasterization, primitive rendering order is important.
Similar to the challenge in 3DGS~\cite{3dgs}, using primitive centers or their closest distance to the camera can produce incorrect ordering, producing popping artifacts~\cite{stopthepop}.
We show two incorrect ordering results using naive sorting criteria in \cref{fig:incorrect_order}.
Thanks to the Octree layout (\cref{sssec:scene_repr}), we can sort by Morton order using our sparse voxels representation.
As illustrated in \cref{fig:cam_quad_and_order}, we can follow certain types of Morton order to render the voxels under an Octree node for correct ordering.
The type of Morton order to follow is solely dependent on the positive/negative signs of the ray direction (the ray origin doesn't matter).
That is to say, we have eight permutations of Morton order for different ray directions in the 3D space.
Finally, the generalization to multi-level voxels can be proved by induction.
An ordering example in 2D with three levels is depicted in \cref{fig:order_example}.

The sorting is applied for each image tile.
In case all the pixels in a tile share the same ray direction signs, we can simply sort the assigned voxels by their type of Morton order.
We handle the corner case when multiple Morton orders are required in supplementary materials.

\paragraph{Rendering pixels.}
Finally, we proceed with alpha composition, \cref{eq:alpha_blend}, to render pixels.
In our case, the $N$ primitives blend of a pixel-ray depends on the number of sparse voxels assigned to the tile that the pixel-ray belongs to.
The computation of the alpha, color, and other geometric properties from our sparse voxels are described in \cref{sssec:scene_repr}.
When rendering sparse voxels for a pixel-ray, we compute ray-aabb intersection to determine the ray segment to sample (for voxel alpha in \cref{eq:vox_alpha}) and skip some non-intersected sparse voxels.
We do early termination of the alpha composition if the transmittance of a sparse voxel is below a threshold $T_i{<}\hypertstop$.


\paragraph{Anti-aliasing.}
To mitigate aliasing artifacts, we render in $\hyperss$ times higher resolution and then apply image downsampling to the target resolution with an anti-aliasing filter.

%% file: sec/3_2_0.tex
\subsection{Progressive Sparse Voxels Optimization}
\label{ssec:pg_sv_optim}
In this section, we describe the procedure to optimize a 3D scene from the input frames with known camera parameters using our \ourshortname presented in \cref{ssec:svraster}.


\paragraph{Voxel max sampling rate.}
We first define the maximum sampling rate $\voxrate$ of each voxel on the training images, which reflects the image region a voxel can cover.
A smaller $\voxrate$ indicates that the voxel is more prone to overfitting due to less observation.
We use $\voxrate$ in our voxel initialization and subdivision process described later.
Given $\ncam$ training cameras, we estimate the maximum sampling rate of a voxel as follows with visualization in \cref{fig:samp_rate}:
\begin{subequations} \label{eq:sampling_rate}
\begin{align}
    \voxrate &= \max\nolimits_{i}^{\ncam} ~\frac{\voxsiz}{\voxinterval^{(i)}} ~, \\
    \voxinterval^{(i)} &= \underbrace{(\voxcen - \rayo^{(i)})^\intercal \ell^{(i)}}_{\text{Voxel z-distance}} \cdot \underbrace{\frac{\tan\left(0.5\cdot\theta_\text{fov-x}^{(i)}\right)}{0.5 \cdot W^{(i)}}}_{\text{Unit-distance pixel size}} ~, \label{eq:sampling_interval}
\end{align}
\end{subequations}
where $\ell$ is camera lookat vector, $\theta_\text{fov-x}$ is camera horizontal field of view, and $W$ is image width. The sampling rate indicates the estimated number of rays along the image's horizontal axis direction that may hit the voxel.

%% file: fig/voxel_init.tex
\begin{figure}
    \centering
    \begin{subfigure}{0.30\linewidth}
        \centering
        \includegraphics[width=\linewidth]{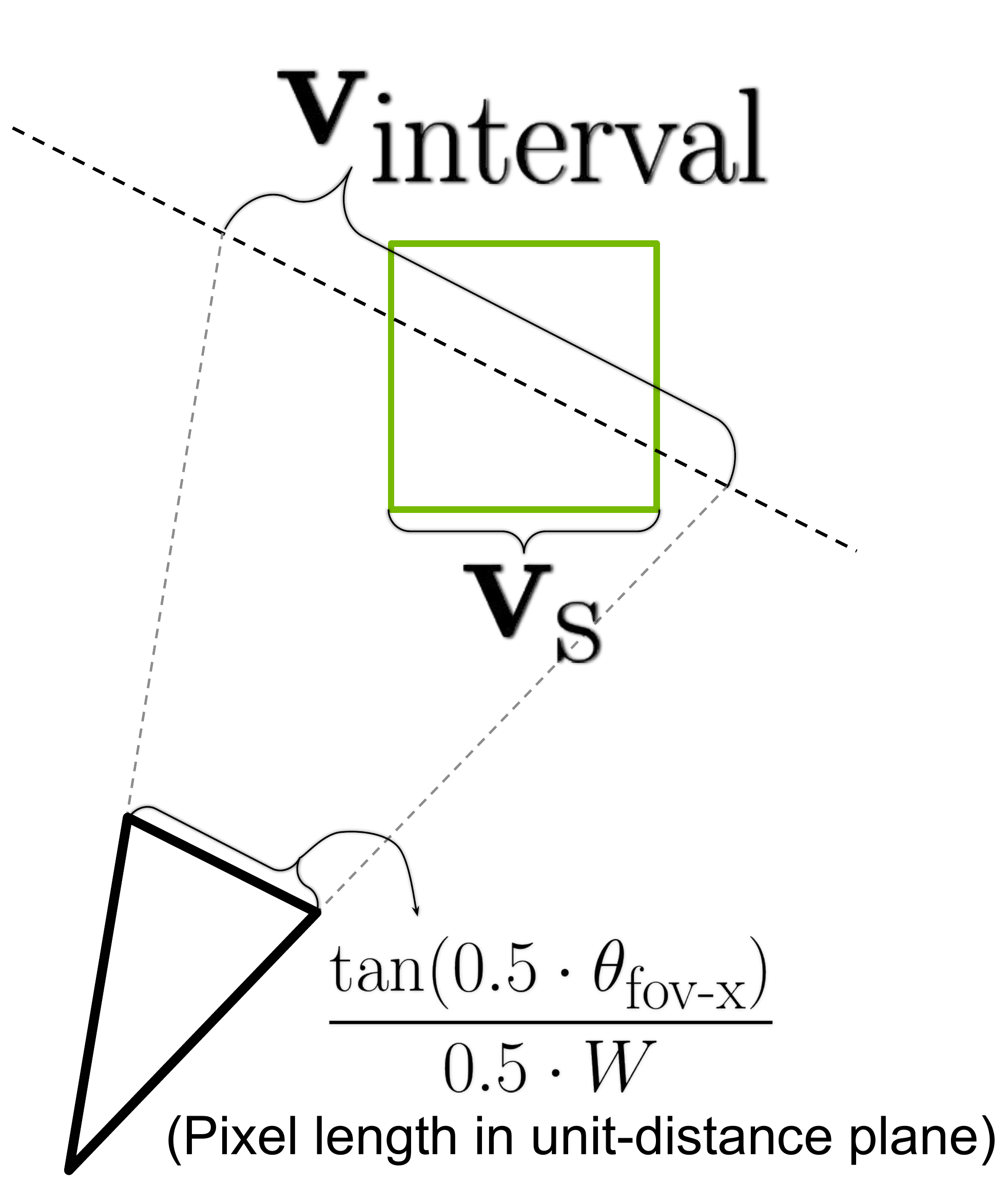}
        \caption{}
        \label{fig:samp_rate}
    \end{subfigure}
    \hfill
    \begin{subfigure}{0.50\linewidth}
        \centering
        \includegraphics[width=\linewidth]{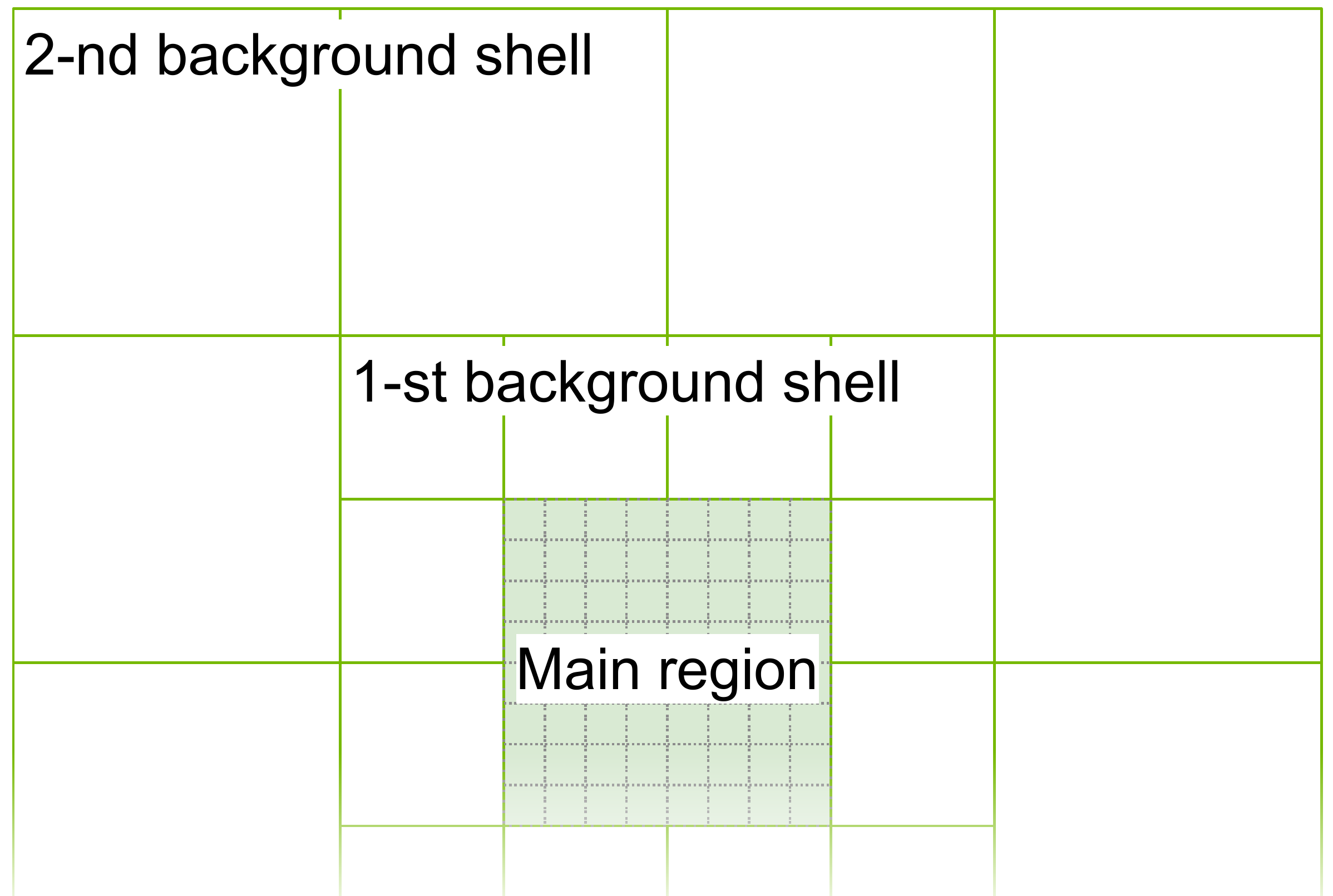}
        \caption{}
        \label{fig:unbounded_init}
    \end{subfigure}
    \hfill
    \caption{
        {\bf Visualization of voxel sampling rate and grid layout initialization.}
        {\bf (a)} We visualize the voxel sampling rate defined in \cref{eq:sampling_rate}.
        {\bf (b)} We depict the foreground main region and the background region under different shell levels. In unbounded scenes, we apply different grid layout initialization strategies for foreground and background regions.
        See \cref{ssec:scene_init} for details.
    }
    \label{fig:grid_init}
    \vspace{-1.5em}
\end{figure}

%% file: sec/3_2_1_scene_init.tex
\subsubsection{Scene Initialization}
\label{ssec:scene_init}
Without employing an additional prior, we initialize all the parameters to constant.
We start with volume density approaching zero by setting voxel raw density to a negative number $\hyperinitgeo$ so that the initial activated density $\explin(\hyperinitgeo){\approx}0$.
We set the SH coefficients to zero for non-zero degrees and set the view-independent zero-degree component to yield gray color (\ie, intensity equal $0.5$).
We detail the Octree grid layout initialization in the following.

\paragraph{Bounded scenes.}
In case the scenes or the objects to reconstruct are enclosed in a known bounded region, we simply initialize the layout as a dense grid with $\hyperinitlv$ levels and remove voxels unobserved by any training images.
The number of voxels is $\le(2^{\hyperinitlv})^3$ after initialization.

\paragraph{Unbounded scenes.}
For the unbounded scenes, we first split the space into the main and the unbounded background regions, depicted in \cref{fig:unbounded_init}, each with a different heuristic.
We use the training camera positions to determine a cuboid for the main region.
The cuboid center is set to the average camera positions and the radius is set to the median distance between the cuboid center and the cameras.
The same as the bounded scenes, we initialize a dense grid with $\hyperinitlv$ levels for the main region.
For the background region, we allocate $\hyperoutlv$ level of background shells enclosing the main region, which means that the radius of the entire scene is $2^{\hyperoutlv}$ of the main region.
In each background shell level, we start with the coarsest voxel size, \ie, $4^3{-}2^3{=}56$ voxels in each shell level.
We then iteratively subdivide shell voxels with the highest sampling rate and remove voxels unobserved by any training cameras.
The process repeats until the ratio of the number of voxels in the background and the main regions is $\hyperoutnumscale$.
The number of voxels is $\le(1+\hyperoutnumscale)(2^{\hyperinitlv})^3$ after initialization.

%% file: sec/3_2_2_scene_adaptive.tex
\subsubsection{Adaptive Pruning and Subdivision}
\label{ssec:scene_adapt}
The initialized grid layout only coarsely covers the entire scene that should be adaptively aligned to different levels-of-detail for the scene during the training progress.
We apply the following two operations every $\hyperevery$ training iterations to achieve this purpose.

\paragraph{Pruning.}
We compute the maximum blending weight ($T_i\alpha_i$) from \cref{eq:alpha_blend} of each voxel using all the training cameras. 
We remove voxels with maximum blending weight lower than $\hyperprune$.

\paragraph{Subdivision.}
Our heuristic is that a voxel with a larger training loss gradient indicates that the voxel region requires finer voxels to model.
Specifically, we accumulate the subdivision priority as the following:
\begin{equation} \label{eq:priority}
    \voxpriority = \sum\nolimits_{\ray\in R} \left\|\alpha(\ray) \cdot \pdv{\loss(\ray)}{\alpha(\ray)}\right\| ~,
\end{equation}
where $R$ is the set of all training pixel rays throughout the $\hyperevery$ iterations and $\loss(\ray)$ is the training loss of the ray.
The gradient is weighted by alpha values contributed from the voxel to the ray.
Higher $\voxpriority$ indicates higher subdivision priority.
To prevent voxels from overfitting few pixels, we set the priority to zero for voxels with maximum sampling rate lower than a sampling rate threshold $\voxrate < 2\hyperrate$.
Finally, we select the voxels with priority above the top $\hyperpercent$ percent to subdivide, \ie, the total number of voxels is increased by $(\hyperpercent \cdot (8-1))$ percent.
Note that we only keep the leaf nodes in the Octree layout so we remove the source voxels once they are subdivided.

When voxels are pruned and subdivided, the voxel Spherical Harmonic (SH) coefficients and the grid point densities need to be updated accordingly.
The SH coefficients are simply pruned together with voxels and duplicated to the subdivided children voxels.
Grid point densities are slightly more complex as the eight voxel corner grid points are shared between adjacent voxels (\cref{fig:scene_repr}).
We remove a grid point only when it does not belong to any voxel corners.
When subdividing, we use trilinear interpolation to compute the densities of the new grid points.
The duplicated grid points are merged and their densities are averaged.

%% file: sec/3_2_3_optim_objective.tex
\subsubsection{Optimization objectives}
\label{sssec:optim_obj}
We use MSE and SSIM as the photometric loss between the rendered and the ground truth images.
The overall training objective is summarized as:
\begin{multline}
    \loss = \loss_{\mathrm{mse}}
    + \lambda_{\mathrm{ssim}} \loss_{\mathrm{ssim}} \\
    + \lambda_{\mathrm{T}} \loss_{\mathrm{T}}
    + \lambda_{\mathrm{dist}} \loss_{\mathrm{dist}}
    + \lambda_{\mathrm{R}} \loss_{\mathrm{R}}
    + \lambda_{\mathrm{tv}} \loss_{\mathrm{tv}} ~,
\end{multline}
where $\lambda$ are the loss weights, $\loss_{\mathrm{T}}$ encourages the final ray transmittances to be either zero or one, $\loss_{\mathrm{dist}}$ is the distortion loss~\cite{mip-nerf360}, $\loss_{\mathrm{R}}$ is the per-point rgb loss~\cite{dvgo}, and $\loss_{\mathrm{tv}}$ is the total variation loss on the sparse density grid.
In mesh extraction task, we also add the depth-normal consistency loss from 2DGS~\cite{2dgs}:
\begin{multline}
    \loss_{\mathrm{mesh}} = \lambda_{\mathrm{n\text{-}dmean}} \loss_{\mathrm{n\text{-}dmean}} + \lambda_{\mathrm{n\text{-}dmed}} \loss_{\mathrm{n\text{-}dmed}} ~,
\end{multline}
where both losses encourage alignment between rendered normals and depth-derived normals from mean and median depth.
More details are in the supplementary materials.

%% file: sec/3_2_4_tsdf_mc.tex
\subsubsection{Sparse-voxel TSDF Fusion and Marching Cubes }
\label{ssec:direct_op}

Our sparse voxels can be seamlessly integrated with grid-based algorithms.
To extract a mesh, we implement Marching Cubes~\cite{marchingcubes} to extract the triangles of an isosurface over density from the sparse voxels.
The duplicated vertices from adjacent voxels are merged to produce a unique set of vertices.
When the adjacent voxels belong to different Octree levels, the extracted triangle may not be connected as the density field is not continuous for voxels in different levels.
Such discontinuities can be removed by simply subdividing all voxels to the finest level.

Deciding the target level set for extracting the isosurface can be tricky for the density field.
Instead, we implement sparse-voxel TSDF-Fusion~\cite{volumefusion,voxelhashing,bundlefusion} to compute the truncated signed distance values of the sparse grid points.
We can then directly extract the surface of the zero-level set using the former sparse-voxel Marching Cubes.
Future extensions of our method could directly model signed distance fields following NeuS~\cite{neus} and VolSDF~\cite{volsdf}.
The sparse-voxel TSDF-Fusion can still be beneficial to directly initialize our sparse voxel representation from sensor depth.

%% file: fig/exp_mipnerf360.tex
\begin{figure}
    \centering
    \includegraphics[width=\linewidth]{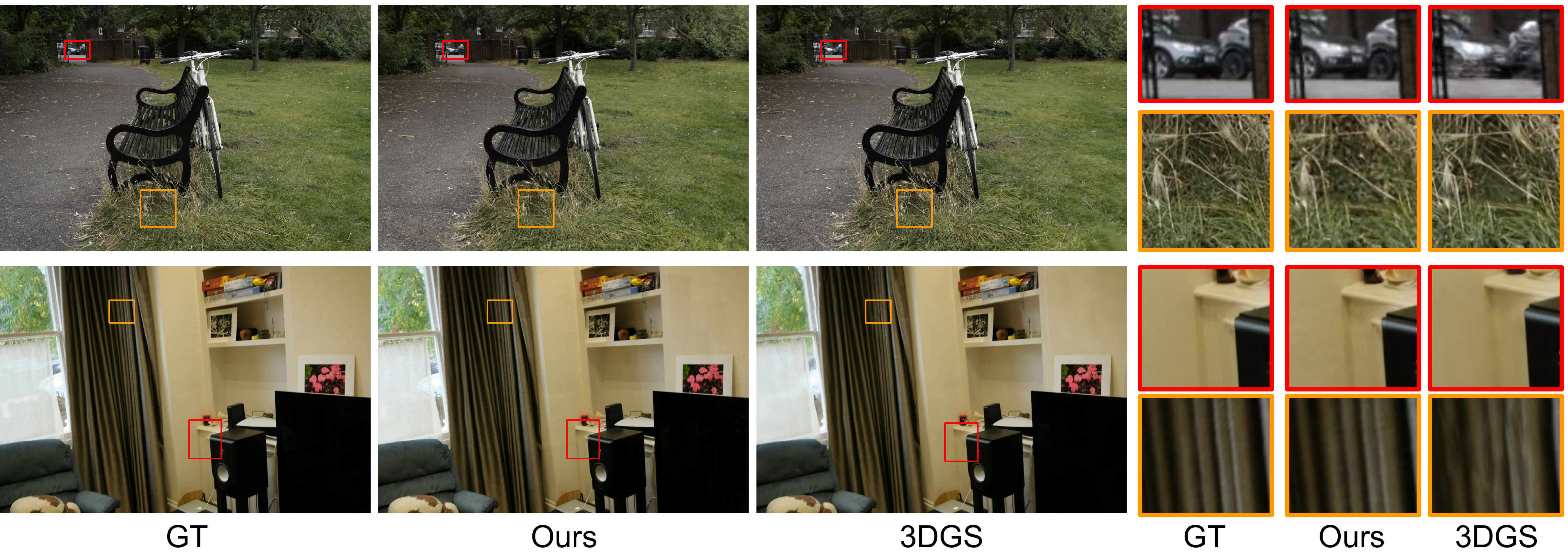}
    \vspace{-1.5em}
    \caption{
    \textbf{A qualitative comparison with 3DGS~\cite{3dgs}.}
    Our result here corresponding to the base version in \cref{tab:mipnerf360}.
    We achieve similar visual quality comparing to 3DGS.
    Note that 3DGS use the coarse geometry from SfM while we do not rely on this prior.
    } 
    \label{fig:qualitative_mipnerf360}
\end{figure}

%% file: fig/exp_meshes.tex
\begin{figure}
    \centering
    \includegraphics[width=\linewidth]{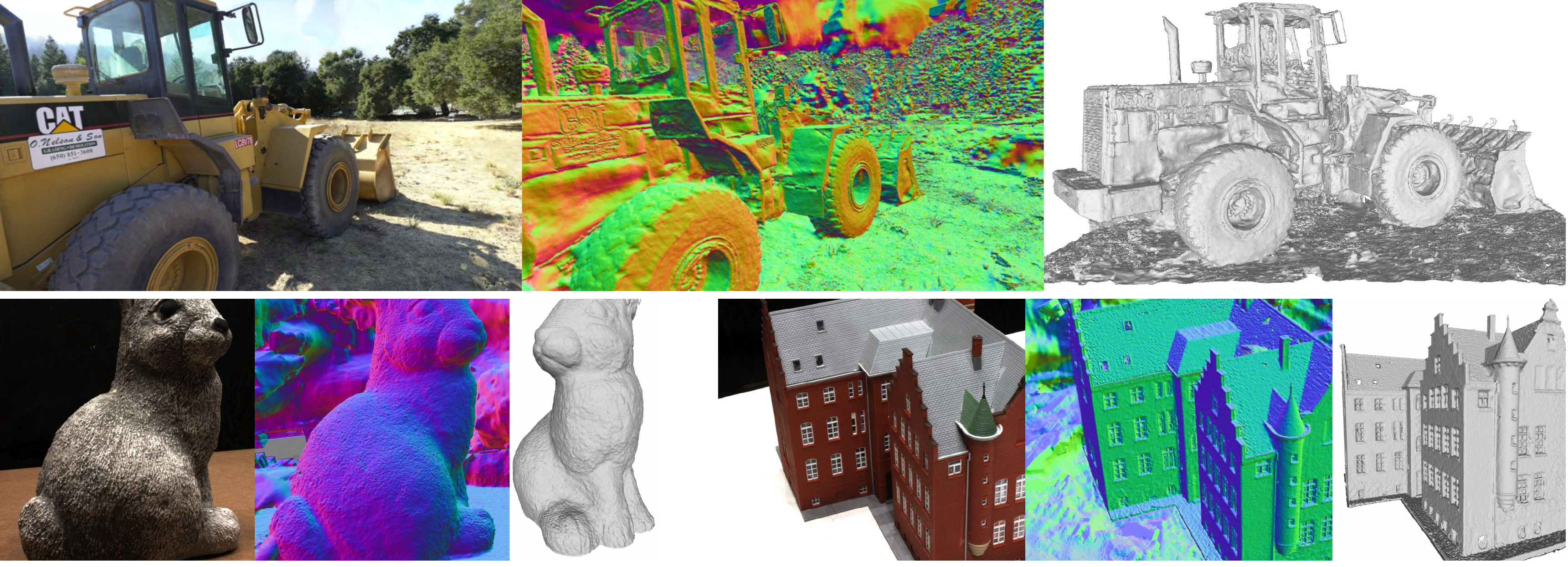}
    \vspace{-1.5em}
    \caption{
    \textbf{Visualization of the reconstructed surface.}
    We show the rendering images, normal maps, and the final meshes on Tanks\&Temples and DTU datasets.
    Note that we only model the scene with density field and do not use the coarse geometry prior from SfM sparse points in this work.
    } 
    \label{fig:qualitative_meshes}
\end{figure}

%% file: tab/mipnerf360.tex
\begin{table}
    \centering
    {\resizebox{1\linewidth}{!}{
    \begin{tabular}{@{}l c@{\hskip 6pt}c c@{\hskip 6pt}c@{\hskip 6pt}c@{}}
        \toprule
        & \multicolumn{5}{c}{Mip-NeRF360 dataset}\\
        \cmidrule{2-6}
        Method & FPS$\uparrow$ & Tr. Time$\downarrow$ & LPIPS$\downarrow$ & PSNR$\uparrow$ & SSIM$\uparrow$ \\
        \midrule
        NeRF~\cite{nerf}
        & $<$1 & $\sim$day & 0.451 & 23.85 & 0.605 \\
        M-NeRF~\cite{mip-nerf}
        & $<$1 & $\sim$day & 0.441 & 24.04 & 0.616 \\
        M-NeRF360~\cite{mip-nerf360}
        & $<$1 & $\sim$day & 0.237 & \silver{27.69} & 0.792 \\
        Zip-NeRF~\cite{zipnerf} \textsuperscript{\textdagger}
        & $<$1 & $\sim$hrs & \silver{0.187} & \gold{28.55} & \gold{0.828} \\
        Plenoxels~\cite{plenoxels}
        & $<$10 & $\sim$30m & 0.463 & 23.08 & 0.626\\
        INGP~\cite{instant-ngp}
        & $\sim$10 & \silver{$\sim$5m} & 0.302 & 25.68 & 0.705\\
        3DGS~\cite{3dgs} \textsuperscript{\textdagger}
        & \silver{131} & 24m & 0.216 & \brown{27.45} & 0.815\\
        {\bf Ours fast-rend} & \gold{258} & \brown{9m} & 0.210 & 26.87 & 0.804\\
        {\bf Ours fast-train} & \silver{131} & \gold{4.5m} & \brown{0.199} & 27.08 & \brown{0.816}\\
        {\bf Ours} & 121 & 15m & \gold{0.185} & 27.33 & \silver{0.822}\\
        \bottomrule
        \multicolumn{6}{@{}l@{}}{\footnotesize {\textsuperscript{\textdagger} Re-evaluate on our machine.}}
    \end{tabular}
    }}
    \vspace{-1em}
    \caption{
        {\bf Novel-view synthesis results comparison on Mip-NeRF360 dataset~\cite{mip-nerf360}.}
        The results are averaged from 4 indoor scenes and 5 outdoor scenes.
        3DGS uses the sparse points prior from COLMAP~\cite{colmap}, whereas the other methods and ours do not.
    }
    \label{tab:mipnerf360}
\end{table}

%% file: tab/tant_db.tex
\begin{table}
    \centering
    {\resizebox{1\linewidth}{!}{
    \begin{tabular}{@{}l c@{\hskip 6pt}c@{\hskip 6pt}c c c@{\hskip 6pt}c@{\hskip 6pt}c}
        \toprule
        & \multicolumn{3}{c}{Tanks\&Temples} && \multicolumn{3}{c}{Deep Blending} \\
        \cmidrule{2-4}\cmidrule{6-8}
        Method & FPS$\uparrow$ & LPIPS$\downarrow$ & PSNR$\uparrow$ &
        & FPS$\uparrow$ & LPIPS$\downarrow$ & PSNR$\uparrow$\\
        \midrule
        Plenoxels~\cite{plenoxels} &
        13 & 0.379 & 21.08 && \brown{11} & 0.510 & 23.06\\
        INGP~\cite{instant-ngp} &
        \brown{14} & 0.305 & 21.92 && 3 & 0.390 & 24.96\\
        M-NeRF360~\cite{mip-nerf360} &
        $<$1 & \brown{0.257} & \brown{22.22} && $<$1 & \brown{0.245} & \brown{29.40}\\
        3DGS~\cite{3dgs} \textsuperscript{\textdagger} &
        \gold{180} & \silver{0.176} & \gold{23.75} && \silver{140} & \silver{0.244} & \silver{29.60}\\
        {\bf Ours} & \silver{125} & \gold{0.144} & \silver{23.04} && \gold{366} & \gold{0.228} & \gold{29.84}\\
        \bottomrule
        \multicolumn{6}{@{}l@{}}{\footnotesize {\textsuperscript{\textdagger} Re-evaluate on our machine.}}
    \end{tabular}
    }}
    \vspace{-1em}
    \caption{
        {\bf Comparison on Tanks\&Temples~\cite{tanksandtemples} and Deep Blending~\cite{deepblending} datasets.}
        We follow 3DGS~\cite{3dgs} to use two outdoor scenes from Tanks\&Temples and two indoor scenes from Deep Blending.
    }
    \label{tab:tant_db}
    \vspace{-.5em}
\end{table}

%% file: fig/failuare_case.tex
\begin{figure}
    \centering
    \includegraphics[width=\linewidth]{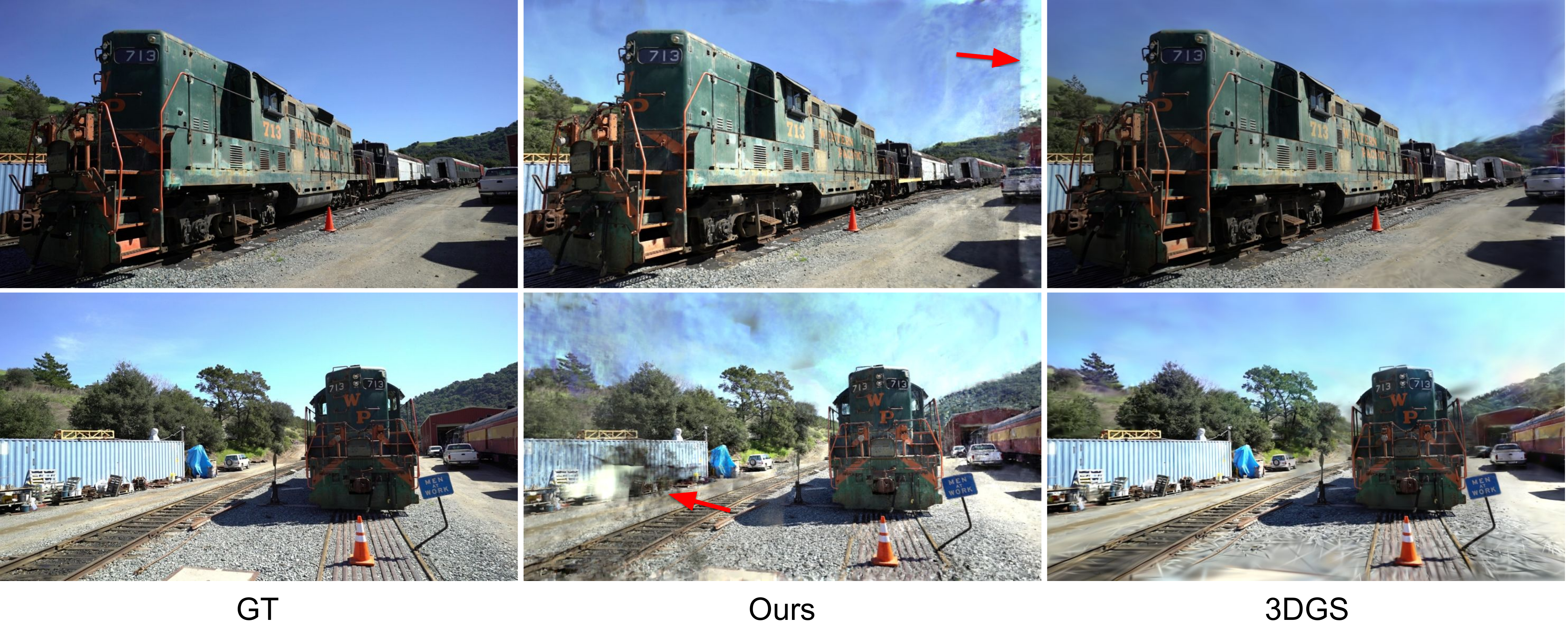}
    \vspace{-1.5em}
    \caption{
    \textbf{Failuare case.}
    On scenes with severe exposure variation of training views, our method struggles and produces clear boundary of different brightnesses and allocates many floaters.
    3DGS on the other hand is less sensitive to photometric variation of GT.
    This explains our worse PSNR and FPS on Tanks\&Temples in \cref{tab:tant_db}.
    } 
    \label{fig:failure_case}
    \vspace{-1.5em}
\end{figure}

%% file: sec/4_0_experiments.tex
\section{Experiments}
\label{sec:experiments}


%% file: sec/4_1_impl_details.tex
\subsection{Implementation Details}
\label{ssec:impl_details}

We use the following implementation details except stated otherwise.
We start the optimization from empty space with raw density set to $\hyperinitgeo{=}{-}10$.
The initial Octree level is $\hyperinitlv{=}6$ (\ie, $64^3$ voxels) for the bounded scenes and the foreground main region of the unbounded scenes.
To model unbounded scenes, we use $\hyperoutlv{=}5$ background shell levels with $\hyperoutnumscale{=}2$ times the number of foreground voxels (\cref{ssec:scene_init,fig:grid_init}).
The early ray stopping threshold is set to $\hypertstop{=}1\mathrm{e}{-4}$ and the supersampling scale is set to $\hyperss{=}1.5$.
Inside each voxel, we sample $\nsamp{=}1$ point for novel-view synthesis and $\nsamp{=}3$ points for mesh reconstruction task.
We train our model for 20K iterations.
The voxels are pruned and subdivided every $\hyperevery{=}1{,}000$ iterations.
The pruning threshold $\hyperprune$ is linearly scale from $0.0001$ to $0.05$ and the subdivision percentage is $\hyperpercent{=}5$.
More details are provided in the supplement.

We use the standard LPIPS~\cite{lpips}, SSIM, and PSNR metrics to evaluate novel-view quality. We directly employ the LPIPS implementation from prior work~\cite{3dgs}, which uses $[0,1]$ image intensity range with VGG~\cite{vgg} network.
For mesh accuracy, we follow the benchmarks to use F-score and chamfer distance.
We use the test-set images to measure the FPS on a desktop computer with a 3090 Ti GPU.

%% file: tab/model_size_scaled_fps.tex
\begin{table}
    \centering
    {\resizebox{1\linewidth}{!}{
    \begin{tabular}{@{}l cc ccc}
        \toprule
        & & & \multicolumn{3}{c}{FPS at higher res.$\uparrow$} \\
        \cmidrule{4-6}
        Method & Peak GPU mem.$\downarrow$ & Model size$\downarrow$ & 1x & 2x & 3x \\
        \midrule
        3DGS & {\bf 1.8GB} & {\bf 0.7GB} & {\bf 131} & 69 & 39 \\
        Ours & 3.9GB & 1.8GB & 121 & {\bf 103} & {\bf 69} \\
        \bottomrule
    \end{tabular}
    }}
    \vspace{-1em}
    \caption{
        {\bf Memory, model size, and high-res FPS.}
        The results are averaged on Mip-NeRF360 dataset.
        The standard 1x evaluation resolution have about 1--1.6M pixels per frame.
    }
    \label{tab:model_size_sacled_fps}
    \vspace{-.5em}
\end{table}

%% file: sec/4_2_nvs.tex
\subsection{Novel-view Synthesis}
\label{ssec:exp_novel_view}


In \cref{tab:mipnerf360}, we show the quantitative comparison on the MipNeRF-360 dataset.
We also provide a fast-rendering variant with \textgreater2x FPS by setting $\hyperss{=}1.5, \hyperprune{=}0.15$, and a fast-training variant with \textgreater3x training speedup by scaling all iteration related hyperparameter by $0.3$.
Both fast variants only reduce quality moderately.

Our rendering speed is comparable to 3DGS on average. However, FPS varies significantly across scenes (see supplementary for detailed per-scene results). Several differences impact speed between our method and 3DGS. For instance, 3DGS sorts 32-bit floats, while we sort 48-bit Morton codes (16 levels, each with 3 bits) for primitive ordering. We also avoid the overhead of computing inverse covariance matrices from quaternions and scaling parameters. We directly intersect rays with voxels in 3D space, whereas 3DGS approximates the projection of 3D Gaussians to 2D via a linear affine transform. Finally, Gaussian distributions decay gradually to zero, while our post-activation~\cite{dvgo,relufield} voxel density field can be arbitrarily sharp. As a result, the average numbers of primitives contributing to a pixel intensity is 63 for 3DGS and 27 for us. The influence of these factors on rendering speed is scene-dependent.

Regarding rendering quality metrics, our method achieves significantly better LPIPS than 3DGS and even surpasses Zip-NeRF. This can be explained by the qualitative comparison in \cref{fig:qualitative_mipnerf360} where our method recover finer details.
The SSIM comparison results with 3DGS is more scene-dependent.
SSIM comparisons with 3DGS are more scene-dependent, while PSNR tends to favor 3DGS, as it prefers smoother rendering in uncertain regions.
More visual comparisons are in the supplementary.

Our method uses much more voxel primitives than Gaussian, resulting in larger model size and requires more GPU memory as shown in \cref{tab:model_size_sacled_fps}.
Interestingly, our high FPS is more scalable to higher resolution rendering perhaps due to the fewer contributing primitives per pixel.

We compare results on two more datasets in \cref{tab:tant_db}.
On Deep Blending~\cite{deepblending} dataset, we achieve much better speed and quality than 3DGS.
On Tanks\&Temples dataset~\cite{tanksandtemples}, despite having better LPIPS, our PSNR and FPS are worse than 3DGS.
The failure results are shown in \cref{fig:failure_case} where the training views have large exposure variation.
As a result, our method produces a clear brightness boundary in the scene with many floaters that slow down our rendering.

%% file: tab/main_abla.tex
\begin{table}
    \centering
    {\resizebox{1\linewidth}{!}{
    \begin{tabular}{@{}lcccc|c@{}}
        \toprule
        Resolution of main & $256^3$ & $512^3$ & $1024^3$ & adaptive & Plenoxels $640^3$ \\
        \midrule
        LPIPS$\downarrow$ & 0.444 & 0.326 & \multirow{3}{*}{OOM} & \gold{0.200} & 0.452\\
        PSNR$\uparrow$ & 23.98 & 25.37 & & \gold{28.01} & 23.29 \\
        FPS$\uparrow$ & \gold{457} & 190 & & 171 & $<$10\\
        \bottomrule
    \end{tabular}
    }}
    \vspace{-1em}
    \caption{
        {\bf Ablation experiments of adaptive and uniform voxel sizes.}
        The results are evaluated on the indoor {\it bonsai} and the outdoor {\it bicycle} scenes from MipNeRF-360 dataset.
        The resolutions at the first row indicate the final grid resolution of the main foreground cuboid.
        Plenxoels is the previous fully-explicit voxel grid approach.
        Please refer to \cref{ssec:exp_ablation} for more discussion.
    }
    \label{tab:main_abla}
    \vspace{-.8em}
\end{table}

%% file: sec/4_3_abla.tex
\subsection{Ablation Studies}
\label{ssec:exp_ablation}

The rasterization process improves the rendering speed of sparse voxels, while the key to achieve high-quality results is the adaptive voxel size for different levels of detail.
We show an ablation experiment in \cref{tab:main_abla}.
Instead of subdividing voxels adaptively, the uniform voxel size variant subdivides all the voxels at certain training iterations until the grid resolution of the foreground main region reaches $256^3$, $512^3$, or $1024^3$.
We also align the background and the pruning setup where the details are deferred to the supplementary.
As shown in the table, the quality of uniform voxel size in $512^3$ resolution is much worse than the adaptive voxel size under similar FPS. Our machine with 24GB GPU memory fails to reach $1024^3$ grid resolution despite the voxels being pruned and sparse.
Plenoxels~\cite{plenoxels} renders sparse voxels by ray casting and sampling following NeRF.
Their FPS is significantly lower than our rasterization-based sparse voxel rendering.
As they use a dense 3D pointer grid to support point sampling on rays, the model scalability for the unbounded background region is thus limited.
To workaround, they apply a multi-sphere image with 64 layers to model the background.
Conversely, we set the scene size as 32x larger ($\hyperoutlv{=}5$) than the foreground cuboid and direct model the entire scene by our sparse voxels.
As a result, our uniform size variant with $512^3$ foreground resolution still outperforms Plenoxels with $640^3$ grid.

In our rasterizer, we use direction-dependent Morton order to ensure the sorting of voxels from different levels is correct.
However, like the case in 3DGS, the popping artifact by the incorrect order is not a major factor in the numerical results.
Instead, we provide more visualization in the supplementary videos to show its effect.
See more ablation studies in the supplementary materials. 

%% file: tab/meshes.tex
\begin{table}[t]
    \centering
    {\resizebox{1\linewidth}{!}{
    \begin{tabular}{@{}l@{\hskip 6pt}c@{\hskip 6pt}c@{\hskip 6pt} c @{\hskip 6pt}c@{\hskip 6pt}c@{\hskip 6pt} c @{\hskip 6pt}c@{\hskip 6pt}c@{}}
    \toprule
    & & && \multicolumn{2}{c}{Tansk\&Temples} && \multicolumn{2}{c}{DTU {\footnotesize}}\\
    \cmidrule{5-6}\cmidrule{8-9}
    Method & Geom. & Init. && F-score$\uparrow$ & Tr. time$\downarrow$ && Cf.$\downarrow$ & Tr. time$\downarrow$\\
    \midrule
    NeRF~\cite{nerf} & Density & Random &&
    - & - && 1.49 & hrs\\
    NeuS~\cite{neus} & SDF & Sphere &&
    \brown{0.38} & hrs && 0.84 & hrs \\
    Voxurf~\cite{voxurf} & SDF & Sphere &&
    - & - && \silver{0.76} & 15m\\
    Neuralangelo~\cite{neuralangelo} & SDF & Sphere &&
    \gold{0.50} & hrs && \gold{0.61} & hrs \\
    3DGS~\cite{3dgs} & Gaussians & SfM pts. &&
    0.19 & \silver{14m} && 1.96 & \silver{11m} \\
    SuGaR~\cite{sugar} & Gaussians & SfM pts. &&
    0.09 & $\sim$1h && 1.33 & $\sim$1h \\
    2DGS~\cite{2dgs} & Gaussians & SfM pts. &&
    0.32 & \brown{16m} && 0.80 & \silver{11m} \\
    {\bf Ours} & Density & Constant &&
    \silver{0.40} & \gold{11m} && \silver{0.76} & \gold{5m}\\
    \bottomrule
    \end{tabular}
    }}
    \vspace{-1em}
    \caption{
    {\bf Mesh results comparison on the Tanks\&Temples~\cite{tanksandtemples} and DTU~\cite{dtu} datasets.}
    Our method with volume density and constant field initialization achieves good accuracy-time trade-off on the unbounded-level and object-level datasets.
    }
    \label{tab:tnt_mesh}
    \vspace{-1em}
\end{table}

%% file: sec/4_4_mesh.tex
\subsection{Mesh Reconstruction}
\label{ssec:exp_mesh_extract}

We extract mesh by adapting TSDF-Fusion~\cite{kinectfusion} and Marching Cubes~\cite{marchingcubes} into our sparse voxels.
The results comparison on the large-scale Tanks\&Temples and the object-scale DTU datasets is provided in \cref{tab:tnt_mesh}, where our method achieves a good balance of accuracy and training time on both datasets.
See qualitative examples in \cref{fig:qualitative_meshes} and the supplementary materials.
We find that our current method tends to produce unnecessary geometric bumps for texture details. Future work could improve surface smoothness by adapting our density field for direct signed distance field modeling or incorporating SfM sparse points prior.

%% file: sec/4_5_feat_fusion.tex
\subsection{2D Feature Fusion}
\label{ssec:feat_fusion}

We showcase some results in \cref{fig:teaser}, where we fuse 2D semantic segmentation and high-dimensional foundation features into our sparse voxels using Voxel Pooling and Volume Fusion.
The multi-view ensemble smooths the inconsistent 2D predictions, while our detailed 3D geometry enables higher-resolution rendering of 2D features.
See more results in the released code.

%% file: sec/5_conclusion.tex
\section{Conclusion}
\label{sec:conclusion}

This work presents a novel differentiable radiance field rendering system that integrates an efficient rasterizer with a multi-level sparse voxel scene representation.
We reconstruct a scene from the multi-view input images by adaptively fitting the sparse voxels into different levels-of-detail.
The results reveal that fully explicit voxel models, without neural networks or Gaussians, can achieve state-of-the-art comparable novel-view rendering speed and quality.
The key breakthrough is overcoming the scalability constraint of voxel grids with adaptive sparse voxels and improving rendering speed through rasterization.
We believe that incorporating our method with classical 3D processing algorithm~\cite{poissonrecon,volumefusion,bundlefusion,voxelhashing,bayesianfusion}
and 3D neural network~\cite{mink,spconv,fVDB} are promising future directions.

%% file: supp_material/main.tex
We provide more details of our method and implementation in \cref{sec:supp_repr_details,sec:supp_rend_details,sec:supp_impl_details}. Some content overlaps with the main paper in the interest of being self-contained. More results and discussion are found in \cref{sec:supp_ablations,sec:supp_results}.

\input{supp_material/detail_repr_0}
\input{supp_material/detail_repr_1_grid}
\input{supp_material/pseudocode/ijk_morton}
\input{supp_material/pseudocode/ray_vox_aabb}
\input{supp_material/detail_repr_2_voxalpha}
\input{supp_material/fig/activation}
\input{supp_material/detail_repr_3_voxnormal}
\input{supp_material/detail_repr_4_voxdepth}
\input{supp_material/detail_rast_0}
\input{supp_material/detail_rast_1_overview}
\input{supp_material/pseudocode/dynamic_morton_order}
\input{supp_material/detail_rast_2_dynamic_morton}
\input{supp_material/detail_rast_3_proof}
\input{supp_material/fig/order_base_case}
\input{supp_material/detail_impl}
\input{supp_material/more_ablation_0}
\input{supp_material/tab/abla_adaptive_vox}
\input{supp_material/more_ablation_1_nvs}
\input{supp_material/more_ablation_2_mesh}
\input{supp_material/tab/abla_nvs_all_in_one}
\input{supp_material/tab/abla_mesh}

\clearpage
\input{supp_material/more_results}


%% file: supp_material/detail_repr_0.tex
\section{More Details of Our Representation}
\label{sec:supp_repr_details}

%% file: supp_material/detail_repr_1_grid.tex
\subsection{Details of Sparse Voxels Grid}
\label{ssec:supp_grid}
Recall that our \ourshortname allocates voxels following an {\bf Octree layout} but does not replicate a traditional {\bf Octree data structure} with parent-child pointers or linear Octree.
We only keep voxels at the Octree leaf nodes without any ancestor nodes and store individual voxels in arbitrary order without the need to maintain a more complex data structure.

The maximum level of detail is set to $\maxlv{=}16$ that defines the finest grid resolution at $65536^3$.
Note that this is only for our CUDA-level implementation convenience.
We leave it as future work to extend to an arbitrary number of levels as we find that $16$ levels are adequate for the scenes we experimented with in this work.

Let $\worldsiz \in \Real$ be the Octree size and $\worldcen \in \Real^3$ be the Octree center in the world space.
The voxel index $v\!=\!\{i,j,k\} \in [0,\ldots,2^{\maxlv}\!\!\!-\!\!1]^3$ together with an Octree level $l\in[1,\maxlv]$ ($l=0$ represent root node and is not used) define voxel size $\voxsiz$ and voxel center $\voxcen$ as:
\begin{align}
    \voxsiz &= \worldsiz \cdot 2^{-l} ~, &
    \voxcen &= \worldcen - 0.5 \cdot \worldsiz + \voxsiz \cdot v ~.
\end{align}
Internally, we map the grid index to its Morton code by a well-known bit interleaving operation, which is helpful to implement our rasterizer detailed later.
A Python pseudocode is provided in \cref{code:conversion_octpath_ijk}.

%% file: supp_material/pseudocode/ijk_morton.tex
\begin{listing}[t]
\begin{minted}
[frame=single,fontsize=\footnotesize]{python}
MAX_NUM_LEVELS = 16

def to_octpath(i, j, k, lv):
    # Input
    #   (i,j,k): voxel index.
    #   lv:      Octree level.
    # Output
    #   octpath: Morton code
    octpath: int = 0
    for n in range(lv):
        bits = 4*(i&1) + 2*(j&1) + (k&1)
        octpath |= bits << (3*n)
        i = i >> 1
        j = j >> 1
        k = k >> 1
    octpath = octpath << (3*(MAX_NUM_LEVELS-lv))
    return octpath

def to_voxel_index(octpath, lv):
    # Input
    #   octpath: Morton code
    #   lv:      Octree level.
    # Output
    #   (i,j,k): voxel index.
    i: int = 0
    j: int = 0
    k: int = 0
    octpath = octpath >> (3*(MAX_NUM_LEVELS-lv))
    for n in range(lv):
        i |= ((octpath&0b100)>>2) << n
        j |= ((octpath&0b010)>>1) << n
        k |= ((octpath&0b001))    << n
        octpath = octpath >> 3
    return (i, j, k)
\end{minted}
\vspace{-1.5em}
\caption{
Pseudocode for conversion between voxel index and Morton code.
See \cref{ssec:supp_grid} for details.
}
\label{code:conversion_octpath_ijk}
\end{listing}

%% file: supp_material/pseudocode/ray_vox_aabb.tex
\begin{listing}[t]
\begin{minted}
[frame=single,fontsize=\footnotesize]{python}
def ray_aabb(vox_c, vox_s, ro, rd):
    # Input
    #   vox_c: Voxel center position.
    #   vox_s: Voxel size.
    #   ro:    Ray origin.
    #   rd:    Ray direction.
    # Output
    #   a:     Ray enter at (ro + a * rd).
    #   b:     Ray exit at (ro + b * rd).
    #   valid: If ray hit the voxel.
    c0 = (vox_c - 0.5 * vox_s - ro) / rd
    c1 = (vox_c + 0.5 * vox_s - ro) / rd
    a = torch.minimum(c0, c1).max()
    b = torch.maximum(c0, c1).min()
    valid = (a <= b) & (a > 0)
    return a, b, valid
\end{minted}
\vspace{-1.5em}
\caption{
Pseudocode for intersecting ray and a axis-aligned voxel.
See \cref{ssec:supp_vox_alpha} for details.
}
\label{code:ray_vox_aabb}
\end{listing}

%% file: supp_material/detail_repr_2_voxalpha.tex
\subsection{Details of Voxel Alpha from Density}
\label{ssec:supp_vox_alpha}

A voxel density field is parameterized by eight parameters attached to its corners $\voxgeo \in \Real^{2\times 2\times 2}$, which is denoted as $\bigv$ for brevity in the later equations.
We use the exponential-linear activation function to map the raw density to non-negative volume density.
We visualize exponential-linear and Softplus in \cref{fig:activation}.
Exponential-linear is similar to Softplus but more efficient to compute on a GPU.
For a sharp density field inside a voxel, we apply the non-linear activation after trilinear interpolation~\cite{dvgo,relufield}.

We evenly sample $\nsamp$ points in the ray segment of ray-voxel intersection to derive the voxel alpha value contributing to the pixel ray.
First, we compute the ray voxel intersection point by \cref{code:ray_vox_aabb}, which yields the ray distances $a$ and $b$ for the entrance and exit points along the ray with ray origin $\rayo\in\Real^3$ and ray direction $\rayd\in\Real^3$.
The coordinate of $k$-th of the K sample points is:
\begin{subequations} \label{eq:K_samp_pts}
\begin{align}
t_k &= a + \frac{k-0.5}{K} \cdot (b - a) \label{eq:ith_depth}\\
\pt_k &= \rayo + t_k \cdot \rayd\\
\rayq_k &= \left(\pt_k - (\voxcen - 0.5 \cdot \voxsiz)\right) \cdot \frac{1}{\voxsiz} ~,
\end{align}
\end{subequations}
where $\pt_k{\in}\Real^3$ is in the world coordinate and $\rayq_k{\in}\Realprob^3$ is in the local voxel coordinate.
The local coordinate $\rayq$ is used to sample voxel by trilinear interpolation:
\begin{gather}
\label{eq:interp}
\interp(\bigv, \rayq) =
  \begin{bmatrix}
{\scriptstyle (1-\rayq_x) \cdot (1-\rayq_y) \cdot (1-\rayq_z)} \\
{\scriptstyle (1-\rayq_x) \cdot (1-\rayq_y) \cdot (\rayq_z)} \\
{\scriptstyle (1-\rayq_x) \cdot (\rayq_y) \cdot (1-\rayq_z)} \\
{\scriptstyle (1-\rayq_x) \cdot (\rayq_y) \cdot (\rayq_z)} \\
{\scriptstyle (\rayq_x) \cdot (1-\rayq_y) \cdot (1-\rayq_z)} \\
{\scriptstyle (\rayq_x) \cdot (1-\rayq_y) \cdot (\rayq_z)} \\
{\scriptstyle (\rayq_x) \cdot (\rayq_y) \cdot (1-\rayq_z)} \\
{\scriptstyle (\rayq_x) \cdot (\rayq_y) \cdot (\rayq_z)}
  \end{bmatrix}^\intercal
  \begin{bmatrix}
   \bigv_{000} \\
   \bigv_{001} \\
   \bigv_{010} \\
   \bigv_{011} \\
   \bigv_{100} \\
   \bigv_{101} \\
   \bigv_{110} \\
   \bigv_{111}
  \end{bmatrix} ~,
\end{gather}
where the subscript in this equation indicates the $x,y,z$ components of the vector $\rayq$ and the sample index is omitted.
Following NeRF~\cite{nerf,Max}, we use quadrature to compute the integrated volume density for alpha value:
\begin{subequations} \label{eq:supp_alpha_comp}
\begin{align}
    \alpha &= 1 - \exp\left( - \frac{\seglen}{\nsamp} \sum_{k=1}^{\nsamp} \explin\left(v_k\right) \right)\\
    v_k &= \interp\left(\bigv, \rayq_k\right) \\
    \seglen &= (b - a) \cdot \|\rayd\| ~,
\end{align}
\end{subequations}
where $\seglen$ is the ray segment length.
The gradient with respect to the voxel density parameters is:
\begin{equation}
    \nabla_{\bigv}~\alpha = (1-\alpha) \cdot \frac{\seglen}{\nsamp} \cdot \sum_{k=1}^K \left( \frac{\dd}{\dd v_k}\explin(v_k) \cdot \nabla_{\bigv} v_k \right) ~.
\end{equation}

%% file: supp_material/fig/activation.tex
\begin{figure}[t]
    \centering
    \includegraphics[width=\linewidth]{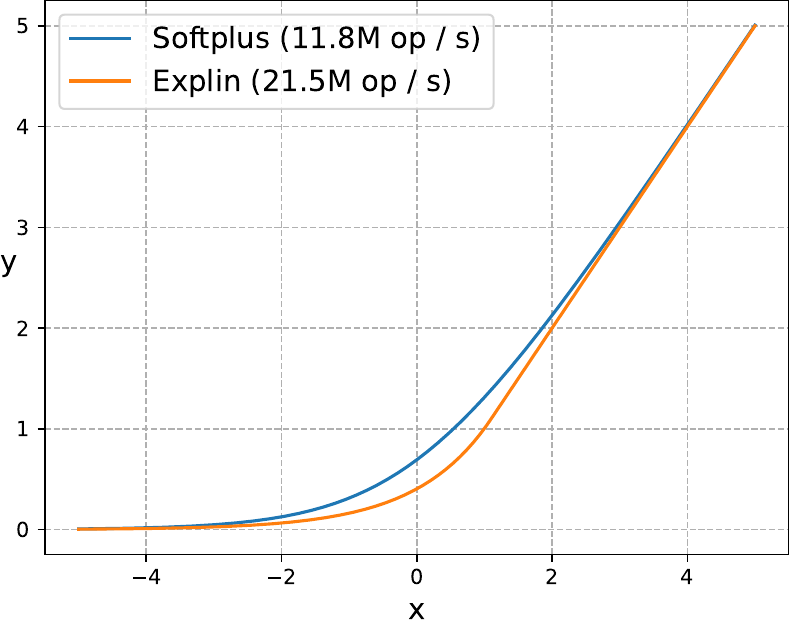}
    \vspace{-2em}
    \caption{
    {\bf Activation functions.}
    We use exponential-linear activation to softly map raw density to non-negative volume density.
    Exp-lin activation is about two times faster to compute in CUDA, which is 21.5M operations per second in a CUDA thread comparing to 11.8M of Softplus.
    See \cref{ssec:supp_vox_alpha} for more details.
    }
    \label{fig:activation}
\end{figure}

%% file: supp_material/detail_repr_3_voxnormal.tex
\subsection{Details of Voxel Normal}
\label{ssec:supp_vox_normal}

Recall that we approximate the normal field as constant inside a voxel for efficiency, which is represented by the analytical gradient of the density field at the voxel center $\rayq^{\mathrm{(c)}}$.
Thanks to the neural-free representation, we derive closed-form equations for forward and backward passes instead of relying on double backpropagation of autodiff.
The unnormalized voxel normal in the forward pass is:
\begin{gather}
\label{eq:interp_grad}
\nabla_{\rayq} \interp(\bigv, \rayq^{\mathrm{(c)}}) = 0.25~\cdot \nonumber\\
  \begin{bmatrix}
   {\scriptstyle (\bigv_{100} + \bigv_{101} + \bigv_{110} + \bigv_{111}) - (\bigv_{000} + \bigv_{001} + \bigv_{010} + \bigv_{011})} \\
   {\scriptstyle (\bigv_{010} + \bigv_{011} + \bigv_{110} + \bigv_{111}) - (\bigv_{000} + \bigv_{001} + \bigv_{100} + \bigv_{101})} \\
   {\scriptstyle (\bigv_{001} + \bigv_{011} + \bigv_{101} + \bigv_{111}) - (\bigv_{000} + \bigv_{010} + \bigv_{100} + \bigv_{110})}
  \end{bmatrix}
\end{gather}
For the backward pass, the gradient with respect to a density parameter is:
\begin{gather}
\label{eq:interp_grad_grad}
\nabla_{\bigv_{ijk}} \nabla_{\rayq} \interp(\bigv, \rayq^{\mathrm{(c)}}) = 0.25 \cdot \begin{bmatrix}
   2i-1 \\
   2j-1 \\
   2k-1
  \end{bmatrix} ~.
\end{gather}

%% file: supp_material/detail_repr_4_voxdepth.tex
\subsection{Details of Voxel Depth}
\label{ssec:supp_vox_depth}

Voxel depths are efficient to compute compared to view-dependent colors and normals so we do the same $K$ points sampling as in the voxel alpha value.
Unlike colors and normals, which are approximated by constant inside each voxel, the depth values of each sample point inside a voxel are different so we need to incorporate the point depth in \cref{eq:ith_depth} into the local alpha composition in \cref{eq:supp_alpha_comp}.
Let
\begin{equation}
    \alpha_k = 1 - \exp\left(-\frac{\seglen}{\nsamp} \cdot \explin\left({\scriptstyle \interp\left(\bigv, \rayq_k\right)}\right) \right)
\end{equation}
be the alpha value of the $k$-th sampled point.
The voxel local depth is:
\begin{equation}
    d = \sum_{k=1}^K \left(\prod_{j=1}^{k-1} (1 - \alpha_j) \right) \cdot \alpha_k \cdot t_k ~.
\end{equation}
Finally, the pixel depth is composited by ${\scriptstyle \mathrm{D}{=}\sum_{i=1}^N T_i d_i}$ from the $N$ voxels, where $T_i$ is the ray transmittance when reaching the $i$-th voxel described in the main paper.

We only experiment with $K\leq 3$ in this work, where the forward and backward equation of each case is summarized as follows.
The $K{=}1$ is trivial with $d{=}\alpha_1 t_1$ and $\frac{\dd d}{\dd \alpha_1}{=}t_1$.
In case $K{=}2$, the backward equations with voxel depth $d{=}\alpha_1 t_1 + (1 - \alpha_1) \alpha_2 t_2$ are:
\begin{equation}
    \frac{\dd d}{\dd \alpha_1} = t_1 - \alpha_2 t_2,~ \quad
    \frac{\dd d}{\dd \alpha_2} = t_2 - \alpha_1 t_2 ~.
\end{equation}
The voxel depth when $K{=}3$ is:
\begin{equation}
    d = \alpha_1 t_1 + (1 - \alpha_1) \alpha_2 t_2 + (1 - \alpha_1) (1 - \alpha_2) \alpha_3 t_3 ~.
\end{equation}
The backward equations are:
\begin{subequations}
\begin{align}
    \frac{\dd d}{\dd \alpha_1} &= t_1 + \alpha_2 \alpha_3 t_3 - \alpha_2 t_2 - \alpha_3 t_3 \\
    \frac{\dd d}{\dd \alpha_2} &= t_2 + \alpha_1 \alpha_3 t_3 - \alpha_1 t_2 - \alpha_3 t_3 \\
    \frac{\dd d}{\dd \alpha_3} &= t_3 + \alpha_1 \alpha_2 t_3 - \alpha_1 t_3 - \alpha_2 t_3 ~.
\end{align}
\end{subequations}

%% file: supp_material/detail_rast_0.tex
\section{More Details of Voxel Rendering Order}
\label{sec:supp_rend_details}

Our sorting-based rasterizer is based on the efficient CUDA implementation done by 3DGS~\cite{3dgs}.
In the following, we first describe how the overall sorting pipeline works in \cref{ssec:supp_sort_overview}.
We then dive more into the implementation of the direction-dependent Morton order in \cref{ssec:supp_sort_morton} and its correctness proof in \cref{ssec:supp_sort_proof}.

A supplementary video is provided to show the effect of correct ordering and a few popping artifacts in comparison with 3DGS~\cite{3dgs}.

%% file: supp_material/detail_rast_1_overview.tex
\subsection{Overview}
\label{ssec:supp_sort_overview}

The goal in the sorting stage of the rasterizer is to arrange a list of voxels in near-to-far order for each image tile.
To this end, 3DGS's rasterizer duplicates a Gaussian for each image tile the Gaussian covers.
A key-value pair is attached to each Gaussian duplication, where the tile index is assigned as the most significant bits of the sorting key.
The bit field of the key-value pair is as follows:
\begin{subequations}
\begin{align}
    \text{key}   &= |\underbrace{\text{tile id}}_{\text{32 bits}}|\underbrace{\text{Gaussian z-depth}}_{\text{32 bits}}|\\
    \text{value} &= |\underbrace{\text{Gaussian id}}_{\text{32 bits}}|
\end{align}
\end{subequations}
By doing so, all the duplicated Guassians assigned to the same image tile will be in the consecutive array segment after sorting with near-to-far z-depth ordering.
In the later rendering stage, each pixel only iterates through the list of Gaussians of its tile for alpha composition.

In our case, we replace the primitive z-depth with a direction-dependent Morton order of voxels to ensure the rendering order is always correct.
As there are eight different Morton orders to follow depending on the positive/negative signs of ray directions, dubbed {\it ray sign bits}, we further duplicate each voxel by the numbers of different ray sign bits it covers.
The ray sign bits are also attached to each duplicated voxel.
In the rendering stage, a pixel only composites voxels with the same attached ray sign bits when there are multiple ray sign bits in an image tile.
Our bit field of the key-value pair is:
\begin{subequations}
\begin{align}
    \text{key}   &= |\underbrace{\text{tile id}}_{\text{16 bits}}|\underbrace{\text{Morton order}}_{\text{48 (=$3\maxlv$) bits}}|\\
    \text{value} &= |\underbrace{\text{ray sign bits}}_{\text{3 bits}}|\underbrace{\text{voxel id}}_{\text{29 bits}}|
\end{align}
\end{subequations}
where $\maxlv{=}16$ is the maximum number of Octree levels.
Note that the ``voxel id" here is indexed to the 1D array location where we store the voxel.
Not to be confused the grid $(i,j,k)$ index in \cref{ssec:supp_grid}.
The bit field arrangement is mainly for our implementation convenient to squeeze everything into 64 and 32 bits unsigned integers.
In our current implementation, the maximum number of tiles is $2^{16}{=}65536$, which is $4096{\times}4096$ maximum image resolution with $16{\times}16$ tile size; the maximum grid resolution is $(2^{16})^3{=}65536^3$; the maximum number of voxels is $2^{29}{\approx}500M$.
We find this is more than enough for the scenes in our experiments.
Future work can define custom data types with extra bits for GPU Radix sort~\cite{radixsort} to increase the resolution limit.

%% file: supp_material/pseudocode/dynamic_morton_order.tex
\begin{listing}[t]
\begin{minted}
[frame=single,fontsize=\footnotesize]{python}
MAX_NUM_LEVELS = 16
order_tables = [
    [0, 1, 2, 3, 4, 5, 6, 7],
    [1, 0, 3, 2, 5, 4, 7, 6],
    [2, 3, 0, 1, 6, 7, 4, 5],
    [3, 2, 1, 0, 7, 6, 5, 4],
    [4, 5, 6, 7, 0, 1, 2, 3],
    [5, 4, 7, 6, 1, 0, 3, 2],
    [6, 7, 4, 5, 2, 3, 0, 1],
    [7, 6, 5, 4, 3, 2, 1, 0],
]

def to_rd_signbits(rd):
    # Input
    #   rd:       Ray direction.
    # Output
    #   signbits: Ray sign bits.
    return 4*(rd[0]<0) + 2*(rd[1]<0) + (rd[2]<0)

def to_dir_dep_morton_order(octpath, signbits):
    # Input
    #   octpath:  Voxel Octree Morton code.
    #   signbits: The signbits the voxel care.
    # Output
    #   order:    The order for sorting.
    table = order_tables[signbits]
    order = 0
    for i in range(MAX_NUM_LEVELS):
        order |= table[octpath & 0b111] << (3*i)
        octpath = octpath >> 3
    return order
\end{minted}
\vspace{-1.5em}
\caption{
Pseudocode for direction-dependent Morton order.
The mapping between voxel grid $(i,j,k)$ index and Octree Morton code \texttt{octpath} is detailed in \cref{code:conversion_octpath_ijk}.
In practice, the mapping from Octree Morton code to direction-dependent Morton order is done by a single bitwise $\mathrm{xor}$ operation instead of for-loop.
More details in \cref{ssec:supp_sort_morton}.
}
\label{code:dynamic_morton_order}
\end{listing}

%% file: supp_material/detail_rast_2_dynamic_morton.tex
\subsection{Direction-dependent Morton Order}
\label{ssec:supp_sort_morton}

As described and illustrated in the main paper, there are eight types of Morton order to follow, each of which is for a certain type of positive/negative signs pattern of ray directions.
We hard-code the eight types of Morton orders, which is used to remap every non-overlapping three bits (corresponding to different Octree levels) in the Octree Morton code of voxels (\cref{ssec:supp_grid}):
\begin{equation}
    ...~b_xb_yb_z~a_xa_ya_z \mapsto ...~f^{\mathrm{(k)}}(b_xb_yb_z)~f^{\mathrm{(k)}}(a_xa_ya_z) ~,
\end{equation}
where $f^{\mathrm{(k)}}: [0\cdots7] \mapsto [0\cdots7]$ is one of the eight permutation mappings.
The pseudocode for computing the ray sign bits and the mapping function from Octree Morton code to direction-dependent Morton order is provided in \cref{code:dynamic_morton_order}.

%% file: supp_material/detail_rast_3_proof.tex
\subsection{Proof of Correct Ordering}
\label{ssec:supp_sort_proof}

We prove the ordering correctness by induction.
We focus on the case for $(+,+,+)$ ray directions.
The proof can be generalized to the other types of ray direction signs by flipping the scene.
The Morton order of the eight voxels in the first Octree level is illustrated in \cref{fig:order_base_case}.

Recap that our sparse voxels only consist of the Octree leaf nodes without any ancestor nodes.
Let $V^{\ell}$ be the space of all valid sparse voxel sets with maximum Octree level equal to $\ell$.
Let $S(\ell)$ be the statement that:
\begin{quote}
    ``For all sparse voxel sets in $V^{\ell}$, their direction-dependent Morton order is always aligned with the near-to-far rendering order for all rays with $(+,+,+)$ direction signs.''
\end{quote}
\paragraph{Base case.}
When $\ell{=}1$, there is only one Octree level.
The direction-dependent Morton order of the eight voxels for $(+,+,+)$ ray directions is illustrated in \cref{fig:order_base_case}.
The bit field from most to least significant bit is for $x$, $y$, and $z$ directions, respectively.
As the ray is going toward $+x$ direction, we can always render the voxels in the $-x$ side ($000, 001, 010, 011$) first before the voxels in the $+x$ side ($100, 101, 110, 111$), which is aligned with the most significant bit of the Morton order.
Similarly, for the voxels in the $-x$ side, we can render the 
voxels in the $-y$ side ($000, 001$) before the $+y$ side ($010, 011$) as the ray is going toward $+y$.
Finally, we can see that the rendering order is correct if we iterate the voxels following the assigned Morton order for ray with $(+,+,+)$ directions.

\paragraph{Induction hypothesis.}
Assume that $S(\ell)$ is true for some positive integer $\ell$.

\paragraph{Induction step.}
We want to show $S(\ell) \implies S(\ell+1)$ is true.
For any sparse voxel set $w \in V^{\ell+1}$, there exists a sparse voxel set $v \in V^{\ell}$ that can evolve into $w$ by: {\it i)} selecting a subset of voxels in $v$ to subdivide with the source voxels removed and {\it ii)} removing some of the voxels.
The $S(\ell)$ indicates that the direction-dependent Morton order of $v$ has the correct rendering order.
To extend for a new Octree level, three zero bits are first append to the least significant bit of the Morton order of every voxel in $v$, which does not affect the ordering.
When subdividing a voxel, the eight child voxels share the same most significant $3\ell$ bits as the source voxel, while the least significant $3$ bits follow the same direction-dependent Morton order as in the base case \cref{fig:order_base_case}.
This reflects the fact that the new child voxels should keep the same relative order to the other voxels as their source parent voxels as the child voxels are all in the 3D space of the source voxels.
The rendering ordering of the eight child voxels can also follow the same Morton order as the base case.
That is the Morton order is still rendering-order correct after subdividing some voxels in $v$.
Finally, removing voxels does not affect the ordering of the remaining others.
In sum, the Morton order of $w$ also has the correct rendering order so $S(\ell)$ implies $S(\ell+1)$.
By induction, $S(\ell)$ is true for all positive integer $\ell$.

%% file: supp_material/fig/order_base_case.tex
\begin{figure}[t]
    \centering
    \includegraphics[width=.6\linewidth]{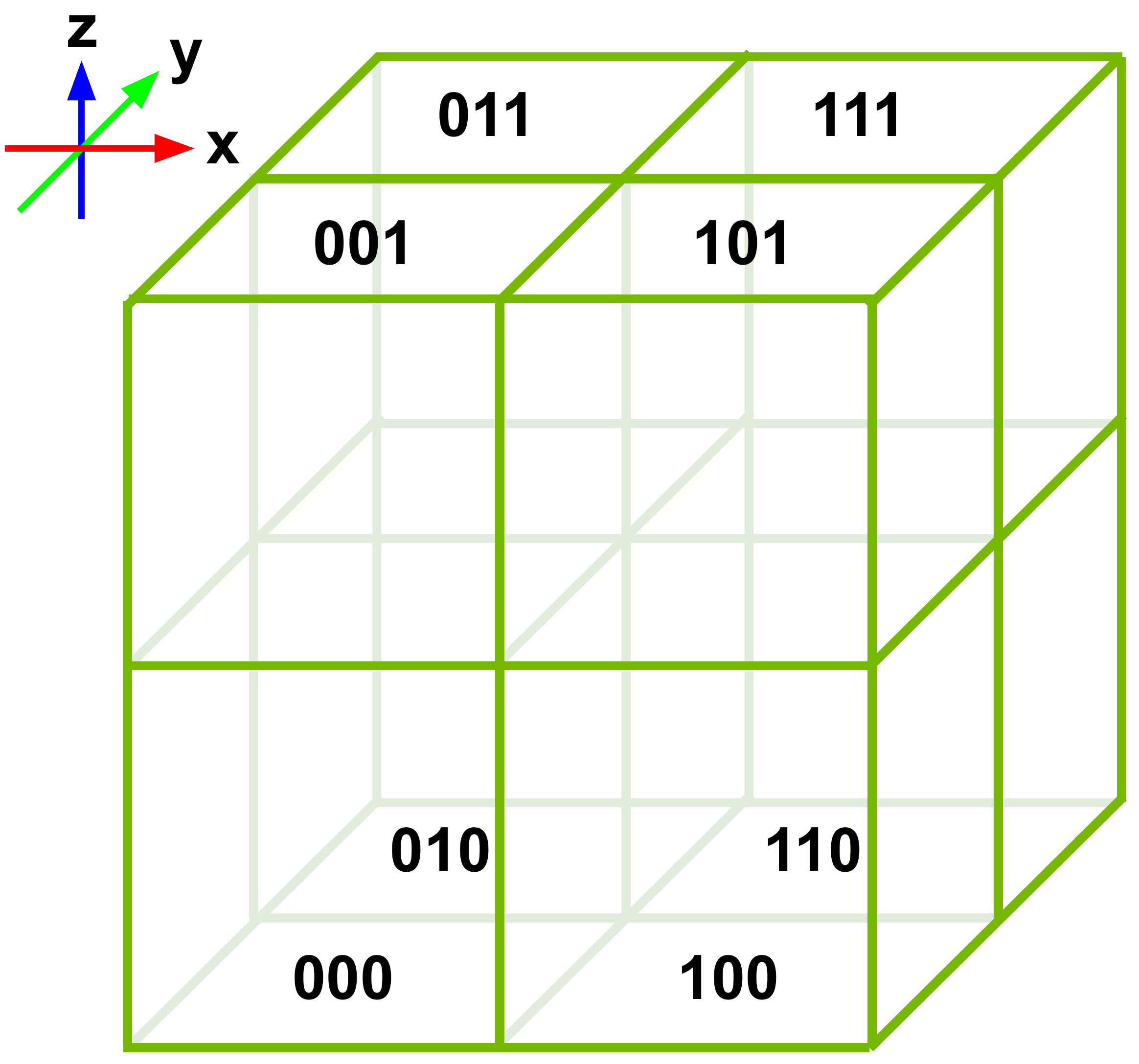}
    \caption{
    {\bf Base case.}
    Direction-dependent Morton order for $(+,+,+)$ ray direction signs under the base case with 1 Octree level.
    The three bits from left to right is for the $x$, $y$, and $z$ directions respectively.
    The rendering order is correct for all rays going toward $(+,+,+)$ direction.
    See \cref{ssec:supp_sort_proof} for more details.
    }
    \label{fig:order_base_case}
\end{figure}

%% file: supp_material/detail_impl.tex
\section{Additional Implementation Details}
\label{sec:supp_impl_details}

We start the optimization from empty space with raw density set to $\hyperinitgeo{=}{-}10$.
We use spherical harmonic (SH) with $\nshd{=}3$ degrees.
The learning rate is set to $0.025$ for the grid point densities, $0.01$ for zero-degree SH coefficients, and $0.00025$ for higher-degree SH coefficients.
We decay all learning rates by $0.1$ at the 19K iteration.
The momentum and the epsilon value of the Adam optimizer are set to $(0.1, 0.99)$ and $1\text{e}{-}15$.
The initial Octree level is $\hyperinitlv{=}6$ (\ie, $64^3$ voxels) for the bounded scenes and the foreground main region of the unbounded scenes.
To model unbounded scenes, we use $\hyperoutlv{=}5$ background shell levels with $\hyperoutnumscale{=}2$ times the number of foreground voxels.
We use average frame color as the color coming from infinite far away for unbounded scenes.
The early ray stopping threshold is set to $\hypertstop{=}1\mathrm{e}{-4}$ and the supersampling scale is set to $\hyperss{=}1.5$.
Inside each voxel, we sample $\nsamp{=}1$ point for novel-view synthesis and $\nsamp{=}3$ points for the mesh reconstruction task.

We train our model for $20K$ iterations.
The voxels are subdivided every $\hyperevery{=}1K$ iterations until $15K$ iterations, where the voxels with top $\hyperpercent{=}5$ percent priority are subdivided each time.
We set $\hyperrate{=}1$ and skip subdividing voxels with a maximum sampling rate below $2\hyperrate$.
The voxels are pruned every $\hyperevery{=}1K$ iterations until $18K$ iterations, where voxels with maximum blending weights less than a pruning threshold are removed.
The pruning threshold is linearly increased from $0.0001$ at the first pruning to $\hyperprune{=}0.05$ at the last pruning.

The loss weights are set to $\lambda_{\mathrm{ssim}}{=}0.02$, $\lambda_{\mathrm{T}}{=}0.01$, $\lambda_{\mathrm{dist}}{=}0.1$ after $10K$ iterations, $\lambda_{\mathrm{R}}{=}0.01$, $\lambda_{\mathrm{tv}}{=}1\text{e}{-}10$ until $10K$ iterations.
For the mesh reconstruction task, the weights for normal-depth alignment self-consistency loss are set to $\lambda_{\mathrm{n\text{-}dmean}}{=}0.001$ and $\lambda_{\mathrm{n\text{-}dmed}}{=}0.001$ for mean and median depth respectively.
The initial depths and normals are bad so the two normal-depth consistency loss is activated at the later training iterations.
We find the median depth converges the fastest so we activate median depth-normal consistency loss at $3K$ iterations, which also only regularizes the rendered depth as median depth is not differentiable.
The mean depth-normal consistency loss is activated at $10K$ iterations.

%% file: supp_material/more_ablation_0.tex
\section{Additional Ablation Studies}
\label{sec:supp_ablations}

%% file: supp_material/tab/abla_adaptive_vox.tex
\begin{table}
    \centering
    \begin{tabular}{@{}lcccc@{}}
        \toprule
        Resolution of main & $256^3$ & $512^3$ & $1024^3$ & adaptive \\
        \midrule
        LPIPS$\downarrow$ & 0.444 & 0.326 & \multirow{3}{*}{OOM} & {\bf 0.200}\\
        PSNR$\uparrow$ & 23.98 & 25.37 & & {\bf 28.01} \\
        FPS$\uparrow$ & {\bf 457} & 190 & & 171\\
        \bottomrule
    \end{tabular}
    \vspace{-.5em}
    \caption{
        {\bf Ablation experiments of adaptive and uniform voxel sizes.}
        The resolutions at the first row indicate the final grid resolution of the main foreground cuboid. Note that OOM is abbreviation of the term, `out-of-memory'.
    }
    \label{tab:supp_abla_adaptive}
\end{table}

%% file: supp_material/more_ablation_1_nvs.tex
\subsection{Novel-View Synthesis}
\label{sec:supp_ablation_nvs}
We conduct comprehensive ablation experiments of our method using the indoor {\bf bonsai} and the outdoor {\bf bicycle} scenes from the MipNeRF-360~\cite{mip-nerf360} dataset.

\paragraph{Adaptive voxel sizes.}
In the main paper, we show that adaptive voxel size for different levels of detail is crucial to achieve high-quality results. The results are recapped in \cref{tab:supp_abla_adaptive}.
We provide experiment details here.
The starting point of the main foreground region is the same for all variants with $64^3$ dense voxels.
Regarding the background region, using the same voxel size as the foreground region is impracticable for uniform-sized variants.
Instead, each of the $5$ background shell voxels is uniformly subdivided by $4$ times as initialization for all the variants.
The difference is that the uniform-sized variants subdivide all voxels each time until the grid resolution of the main region reaches $256^3$, $512^3$, or $1024^3$ instead of subdividing voxels adaptively as described in the main paper.
The pruning setup remains the same for all variants.
The result in \cref{tab:supp_abla_adaptive} shows that adaptive voxel sizes are the key to solve the scalability issue of uniform-sized voxel, which achieves much better rendering quality with high render FPS.

\paragraph{More ablation studies for the hyperparameters.}
We conduct more ablation experiments to show the effectiveness of the hyperparameters in \cref{tab:supp_abla_K,tab:supp_abla_shd,tab:supp_abla_shd,tab:supp_abla_ss,tab:supp_abla_prune,tab:supp_abla_subdiv,tab:supp_abla_outrate,tab:supp_nd_loss,tab:supp_dist_loss,tab:supp_rgbcon_loss,tab:supp_tv_loss,tab:supp_Tcon_loss}.
We mark the adopted hyperparameter setup by ``*" in the table rows.
The setup of the marked rows across different tables can be different as we update the base setups in a rolling manner during the hyperparameter tuning stage.
In each table, the other hyperparameter setups except the ablated one are the same.
We discuss the experiments directly in the table captions to avoid the need for cross-referencing between the tables and the main text.

%% file: supp_material/more_ablation_2_mesh.tex
\subsection{Mesh Reconstruction}
\label{sec:supp_ablation_mesh}
To show the effectiveness of the mesh regularization losses, we use the {\bf Ignatius} and the {\bf Truck} scenes from TnT~\cite{tanksandtemples} dataset and three scans with id {\bf 24, 69, 122} from DTU~\cite{dtu} dataset for ablation studies.
The results are shown in \cref{tab:supp_mesh_reg}.
While the normal-depth self-consistency losses do not improve novel-view synthesis quality in \cref{tab:supp_nd_loss}, the mesh accuracy is improved by an obvious margin with the regularizations.

%% file: supp_material/tab/abla_nvs_all_in_one.tex
\begin{table}[h]
    \centering
    {\resizebox{1\linewidth}{!}{
    \begin{tabular}{@{}l c c c c c@{}}
        \toprule
        Setup & FPS$\uparrow$ & Tr. time$\downarrow$ & LPIPS$\downarrow$ & PSNR$\uparrow$ & SSIM$\uparrow$ \\
        \midrule
        no ss              & {\bf 111} &    {\bf 13.5m} &    0.201 &   27.69 &   0.830 \\
$\hyperss{=}1.01$  & 108 &    {\bf 13.5m} &    0.193 &   28.24 &   0.845 \\
$\hyperss{=}1.10$* & 107 &    {\bf 13.5m} &    0.190 &   28.32 &   0.848 \\
$\hyperss{=}1.20$  &  99 &    13.6m &    0.188 &   28.36 &   0.849 \\
$\hyperss{=}1.30$  & 100 &    13.7m &    0.187 &   28.39 &   0.850 \\
$\hyperss{=}1.50$  &  92 &    13.8m &    0.186 &   28.42 &   0.851 \\
$\hyperss{=}2.00$  &  75 &    14.2m &    {\bf 0.185} &   {\bf 28.46} &   {\bf 0.853} \\
        \bottomrule
    \end{tabular}
    }}
    \vspace{-.5em}
    \caption{
        {\bf Supersampling rate.}
        Our rendering suffers from aliasing artifact so we render the image in $\hyperss\times$ higher resolution and apply image downsampling with anti-aliasing filter.
        The quality without supersampling is much worse than the others.
        Resampling the image with a very small $\hyperss=1.01$ can already boost quality significantly.
        We find the quality can keep going better with higher $\hyperss$ but the FPS drops by more than $30\%$ at $\hyperss=2$.
        More future development is needed for a more efficient anti-aliasing rendering of our method.
        We use $\hyperss=1.1$ for speed-quality trade-off. 
    }
    \label{tab:supp_abla_ss}
\end{table}

\begin{table}[h]
    \centering
    {\resizebox{1\linewidth}{!}{
    \begin{tabular}{@{}l c c c c c@{}}
        \toprule
        Setup & FPS$\uparrow$ & Tr. time$\downarrow$ & LPIPS$\downarrow$ & PSNR$\uparrow$ & SSIM$\uparrow$ \\
        \midrule
$\nshd{=}1$  & {\bf 118} &    {\bf 11.2m} &    0.201 &   27.43 &   0.840 \\
$\nshd{=}2$  & 114 &    12.1m &    0.193 &   27.94 &   0.847 \\
$\nshd{=}3$* & 107 &    13.5m &    {\bf 0.190} &   {\bf 28.32} &   {\bf 0.848} \\
        \bottomrule
    \end{tabular}
    }}
    \vspace{-.5em}
    \caption{
        {\bf Degree of Spherical Harmonic (SH).}
        The rendering time with higher SH degree is similar but the quality is much better. We use $\nshd{=}3$ as our final setup.
        However, about $80\%$ of the parameters and the disk space is occupied by the SH coefficient with $\nshd{=}3$.
        Future work may want to design a more parameters efficient representation for view-dependent colors.
    }
    \label{tab:supp_abla_shd}
\end{table}

\begin{table}[h]
    \centering
    {\resizebox{1\linewidth}{!}{
    \begin{tabular}{@{}l c c c c c@{}}
        \toprule
        Setup & FPS$\uparrow$ & Tr. time$\downarrow$ & LPIPS$\downarrow$ & PSNR$\uparrow$ & SSIM$\uparrow$ \\
        \midrule
$K{=}1$* & {\bf 107} &    {\bf 13.5m} &    0.190 &   28.32 &   0.848 \\
$K{=}2$  & 102 &    13.8m &    {\bf 0.189} &   28.32 &   {\bf 0.849} \\
$K{=}3$  &  99 &    13.9m &    {\bf 0.189} &   {\bf 28.33} &   {\bf 0.849} \\
        \bottomrule
    \end{tabular}
    }}
    \includegraphics[width=.9\linewidth]{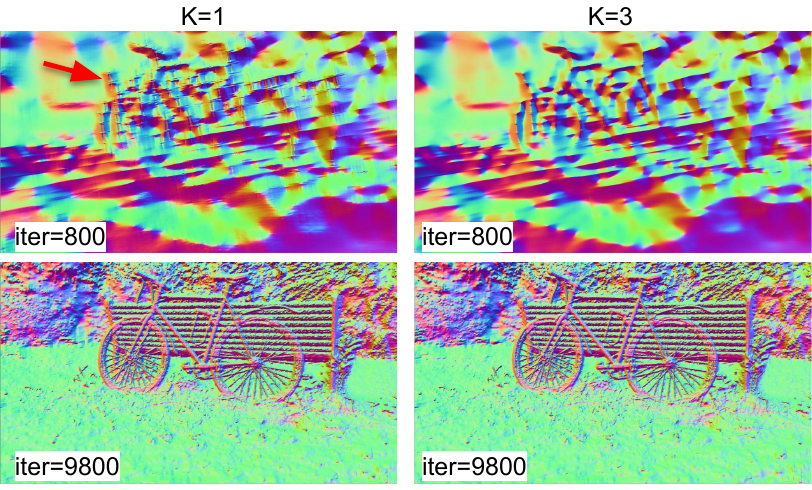}
    \vspace{-.5em}
    \caption{
        {\bf Number of sample points in a voxel when rendering.}
        The effect of sampling more point inside a voxel is marginal as the voxels are typically subdivided into fine level with small size.
        It mainly affects the depth rendering for larger voxels.
        The figure shows the normal derived from the rendered depth.
        $K{=}1$ at the early training stage produce noisy depth as highlighted by the red arrow, while the depth noisy is mitigated when the voxels is subdivided into finer level.
        We suggest to use $K{>}1$ only when the sub voxel depth accuracy is required.
    }
    \label{tab:supp_abla_K}
\end{table}

\begin{table}[h]
    \centering
    {\resizebox{1\linewidth}{!}{
    \begin{tabular}{@{}l c c c c c@{}}
        \toprule
        Setup & FPS$\uparrow$ & Tr. time$\downarrow$ & LPIPS$\downarrow$ & PSNR$\uparrow$ & SSIM$\uparrow$ \\
        \midrule
$\hyperprune{=}0.01$  &  96 &    16.7m &    0.191 &   {\bf 28.41} &   0.847 \\
$\hyperprune{=}0.03$  & 107 &    13.5m &    {\bf 0.190} &   28.32 &   {\bf 0.848} \\
$\hyperprune{=}0.05$* & 119 &    11.6m &    0.192 &   28.20 &   0.846 \\
$\hyperprune{=}0.10$  & 158 &     9.1m &    0.199 &   27.95 &   0.840 \\
$\hyperprune{=}0.15$  & 188 &     7.7m &    0.212 &   27.72 &   0.831 \\
$\hyperprune{=}0.20$  & 213 &     6.8m &    0.224 &   27.53 &   0.823 \\
$\hyperprune{=}0.30$  & {\bf 241} &     {\bf 5.9m} &    0.248 &   27.22 &   0.806 \\
        \bottomrule
    \end{tabular}
    }}
    \vspace{-.5em}
    \caption{
        {\bf Pruning threshold.}
        We prune voxels with maximum blending weights below $\hyperprune$.
        Higher FPS and faster processing time can be achieved by pruning more voxels but with loss in quality.
        We finally use $\hyperprune{=}0.05$ to balance speed and quality.
    }
    \label{tab:supp_abla_prune}
\end{table}

\begin{table}[h]
    \centering
    {\resizebox{1\linewidth}{!}{
    \begin{tabular}{@{}l c c c c c@{}}
        \toprule
        Setup & FPS$\uparrow$ & Tr. time$\downarrow$ & LPIPS$\downarrow$ & PSNR$\uparrow$ & SSIM$\uparrow$ \\
        \midrule
$0.33$ & {\bf 159} &     {\bf 9.9m} &    0.210 &   27.98 &   0.833 \\
$0.50$ & 130 &    11.2m &    0.197 &   28.16 &   0.843 \\
$1.00$*& 107 &    13.5m &    {\bf 0.190} &   28.32 &   {\bf 0.848} \\
$2.00$ & 101 &    15.0m &    {\bf 0.190} &   {\bf 28.36} &   {\bf 0.848} \\
$3.00$ & 101 &    15.7m &    {\bf 0.190} &   {\bf 28.36} &   {\bf 0.848} \\
        \bottomrule
    \end{tabular}
    }}
    \vspace{-.5em}
    \caption{
        {\bf Subdivision scale.}
        We subdivide $\hyperpercent{=}5$ percent of the voxels with the highest priority 15 times during the training.
        As the number of voxels become $(1 + 0.07\hyperpercent)$ at each subdivision, the subdivision scales in above table shows their $\frac{(1 + 0.07\hyperpercent)^{15}}{1.35^{15}}$, which indicate the expected relative number of voxels comparing to the base setup.
        The merit of subdividing more voxels each time is marginal comparing to the base setup.
    }
    \label{tab:supp_abla_subdiv}
\end{table}

\begin{table}[h]
    \centering
    {\resizebox{1\linewidth}{!}{
    \begin{tabular}{@{}l c c c c c@{}}
        \toprule
        Setup & FPS$\uparrow$ & Tr. time$\downarrow$ & LPIPS$\downarrow$ & PSNR$\uparrow$ & SSIM$\uparrow$ \\
        \midrule
$\hyperoutnumscale{=}1.0$ & {\bf 120} &    {\bf 11.6m} &    0.195 &   28.17 &   0.844 \\
$\hyperoutnumscale{=}2.0$*& 107 &    13.5m &    0.190 &   28.32 &   0.848 \\
$\hyperoutnumscale{=}3.0$ & 103 &    14.7m &    {\bf 0.189} &   28.35 &   {\bf 0.849} \\
$\hyperoutnumscale{=}4.0$ & 103 &    15.5m &    {\bf 0.189} &   {\bf 28.38} &   {\bf 0.849} \\
        \bottomrule
    \end{tabular}
    }}
    \vspace{-.5em}
    \caption{
        {\bf Initial ratio of the number of voxels in background and main regions.}
        At the initialization stage, we heuristically subdivide voxel in the background region until the ratio of the number of voxel is $\hyperoutnumscale$ to the foreground region.
        The overall result quality are similar for different $\hyperoutnumscale$.
        It affects training time more than testing FPS as the training iterations per second before any pruning is depend on the initial number of voxels.
    }
    \label{tab:supp_abla_outrate}
\end{table}

\begin{table}[h]
    \centering
    {\resizebox{1\linewidth}{!}{
    \begin{tabular}{@{}l c c c c c@{}}
        \toprule
        Setup & FPS$\uparrow$ & Tr. time$\downarrow$ & LPIPS$\downarrow$ & PSNR$\uparrow$ & SSIM$\uparrow$ \\
        \midrule
$\lambda_{\mathrm{tv}}{=}0$ & 102 &    14.0m &    0.202 &   27.77 &   0.832 \\
$\lambda_{\mathrm{tv}}{=}1\text{e}{-}11$ & 106 &    13.6m &    0.196 &   27.97 &   0.840 \\
$\lambda_{\mathrm{tv}}{=}1\text{e}{-}10$*& {\bf 107} &    {\bf 13.5m} &    {\bf 0.190} &   {\bf 28.32} &   {\bf 0.848} \\
$\lambda_{\mathrm{tv}}{=}1\text{e}{-}9$ &  99 &    15.5m &    0.213 &   27.85 &   0.822 \\
        \bottomrule
    \end{tabular}
    }}
    \includegraphics[width=\linewidth]{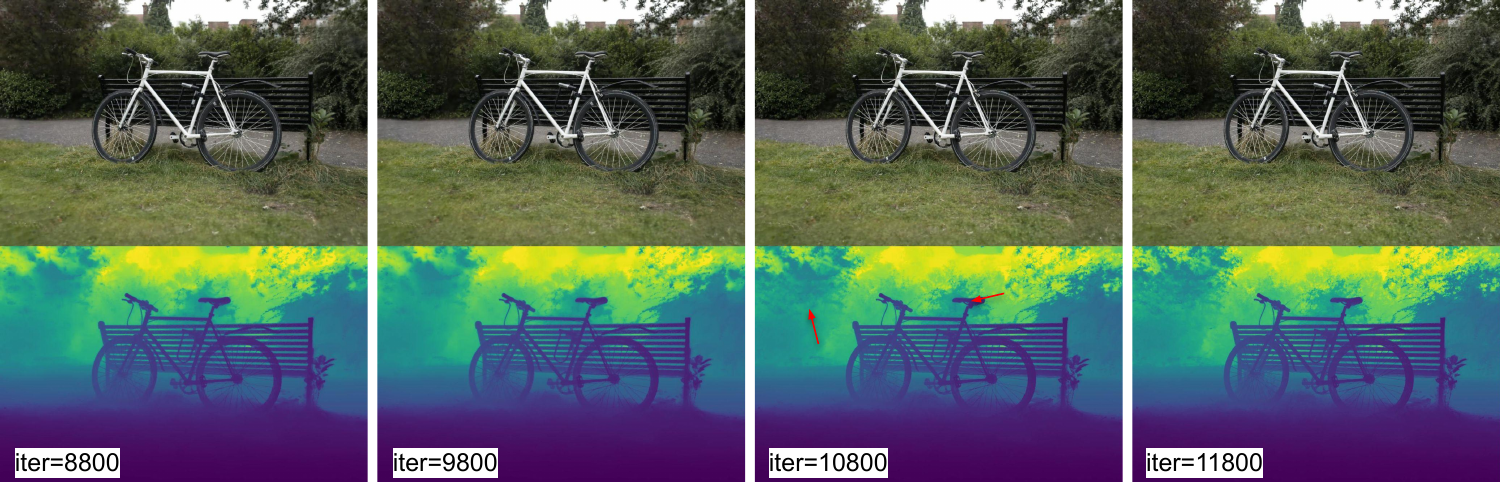}
    \vspace{-.5em}
    \caption{
        {\bf Total Variation (TV) loss.}
        Similar to previous grid-based approaches~\cite{dvgo,plenoxels,tensorf}, TV loss is also important in our method.
        We apply TV loss on density grid only for the first half $10{,}000$ iterations as applying TV for all iterations leads to blurrier rendering.
        TV with proper loss weighting leads to better quantitative results without loss of speed.
        The effect of TV loss is also visualized in above figure, where many geometric details emerge after the TV loss is turned off.
        The employed TV loss scheduling entails the coarse-to-fine optimization strategy.
    }
    \label{tab:supp_tv_loss}
\end{table}

\begin{table}[h]
    \centering
    {\resizebox{1\linewidth}{!}{
    \begin{tabular}{@{}l c c c c c@{}}
        \toprule
        Setup & FPS$\uparrow$ & Tr. time$\downarrow$ & LPIPS$\downarrow$ & PSNR$\uparrow$ & SSIM$\uparrow$ \\
        \midrule
$\lambda_{\mathrm{R}}{=}0$             & 111 &    14.6m &    0.200 &   28.11 &   0.843 \\
$\lambda_{\mathrm{R}}{=}1\text{e}{-}4$ & 110 &    14.6m &    0.199 &   28.11 &   0.843 \\
$\lambda_{\mathrm{R}}{=}1\text{e}{-}3$ & 107 &    14.5m &    0.196 &   28.20 &   0.845 \\
$\lambda_{\mathrm{R}}{=}1\text{e}{-}2$*& 107 &    13.5m &    {\bf 0.190} &   {\bf 28.32} &   {\bf 0.848} \\
$\lambda_{\mathrm{R}}{=}1\text{e}{-}1$ & {\bf 118} &    {\bf 10.9m} &    0.205 &   27.77 &   0.830 \\
        \bottomrule
    \end{tabular}
    }}
    \vspace{-.5em}
    \caption{
        {\bf Color concentration loss.}
        We find it helpful to apply L2 loss directly between observed pixel color and the individual voxel color of each voxel passing by the ray~\cite{dvgo}, which slightly improve training time and result quality.
    }
    \label{tab:supp_rgbcon_loss}
\end{table}

\begin{table}[h]
    \centering
    {\resizebox{1\linewidth}{!}{
    \begin{tabular}{@{}l c c c c c@{}}
        \toprule
        Setup & FPS$\uparrow$ & Tr. time$\downarrow$ & LPIPS$\downarrow$ & PSNR$\uparrow$ & SSIM$\uparrow$ \\
        \midrule
$\lambda_{\mathrm{dist}}{=}0$                      & 105 &    14.9m &    0.199 &   27.27 &   0.839 \\
$\lambda_{\mathrm{dist}}{=}1\text{e}{-}4$          & 106 &    15.4m &    0.199 &   27.48 &   0.839 \\
$\lambda_{\mathrm{dist}}{=}1\text{e}{-}3$          & 105 &    15.1m &    0.195 &   27.97 &   0.842 \\
$\lambda_{\mathrm{dist}}{=}1\text{e}{-}2$          & 107 &    13.5m &    0.190 &   {\bf 28.32} &   {\bf 0.848} \\
$\lambda_{\mathrm{dist}}{=}1\text{e}{-}1$          & {\bf 137} &     {\bf 9.9m} &    0.256 &   26.34 &   0.760 \\
$\lambda_{\mathrm{dist}}{=}1\text{e}{-}4$ from 10K & 104 &    15.1m &    0.199 &   27.40 &   0.839 \\
$\lambda_{\mathrm{dist}}{=}1\text{e}{-}3$ from 10K & 105 &    15.0m &    0.197 &   27.76 &   0.842 \\
$\lambda_{\mathrm{dist}}{=}1\text{e}{-}2$ from 10K & 105 &    14.6m &    0.193 &   28.08 &   0.845 \\
$\lambda_{\mathrm{dist}}{=}1\text{e}{-}1$ from 10K*& 113 &    13.7m &    {\bf 0.188} &   28.11 &   {\bf 0.848} \\
        \bottomrule
    \end{tabular}
    }}
    \includegraphics[width=.9\linewidth]{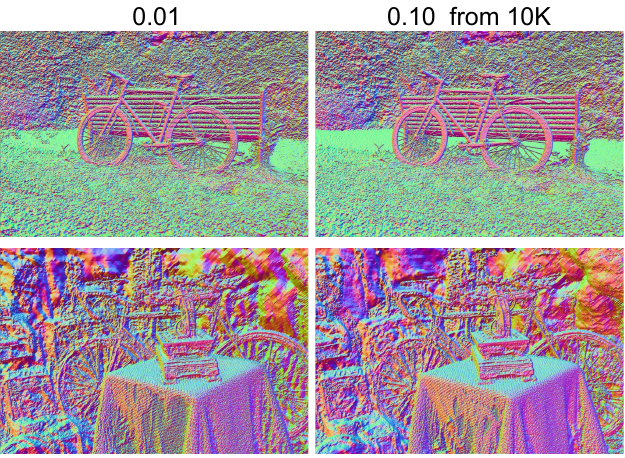}
    \vspace{-.5em}
    \caption{
        {\bf Distortion loss.}
        Distortion loss is proposed by MipNeRF-360~\cite{mip-nerf360} and employed by many NeRF-based rendering approaches to encourage concentration of the blending weight distribution on a ray.
        We find distortion loss is also helpful in our method, especially for the PSNR.
        We also find that employing a larger distortion loss weight after the total variation loss is turned off lead to a cleaner geometry as shown in the above depth-derived normal visualization.
    }
    \label{tab:supp_dist_loss}
\end{table}

\begin{table}[h]
    \centering
    {\resizebox{1\linewidth}{!}{
    \begin{tabular}{@{}l c c c c c@{}}
        \toprule
        Setup & FPS$\uparrow$ & Tr. time$\downarrow$ & LPIPS$\downarrow$ & PSNR$\uparrow$ & SSIM$\uparrow$ \\
        \midrule
$\lambda_{\mathrm{T}}{=}1\text{e}{-}0$ & {\bf 109} &    13.4m &    0.192 &   28.23 &   0.847 \\
$\lambda_{\mathrm{T}}{=}1\text{e}{-}3$ & {\bf 109} &    13.5m &    0.191 &   28.25 &   0.847 \\
$\lambda_{\mathrm{T}}{=}1\text{e}{-}2$*& 107 &    13.5m &    {\bf 0.190} &   {\bf 28.32} &   {\bf 0.848} \\
$\lambda_{\mathrm{T}}{=}1\text{e}{-}1$ & {\bf 109} &    {\bf 12.8m} &    0.192 &   28.13 &   0.845 \\
        \bottomrule
    \end{tabular}
    }}
    \vspace{-.5em}
    \caption{
        {\bf Transmittance concentration loss.}
        The effect of encouraging final ray transmittance to be either zero or one is marginal in the unbounded scenes.
        We find this loss is more important for the object-centric scenes with foreground region only and known background colors (\eg, Synthetic-NeRF dataset~\cite{nerf}).
    }
    \label{tab:supp_Tcon_loss}
\end{table}

\begin{table}[h]
    \centering
    {\resizebox{1\linewidth}{!}{
    \begin{tabular}{@{}l c c c c c@{}}
        \toprule
        Setup & FPS$\uparrow$ & Tr. time$\downarrow$ & LPIPS$\downarrow$ & PSNR$\uparrow$ & SSIM$\uparrow$ \\
        \midrule
neither*    & 107 &    {\bf 13.5m} &    {\bf 0.190} &   {\bf 28.32} &   0.848 \\
n-dmed  & {\bf 114} &    {\bf 13.5m} &    0.191 &   28.10 &   0.849 \\
n-dmean &  97 &    14.0m &    {\bf 0.190} &   28.14 &   {\bf 0.850} \\
both    & 103 &    14.4m &    0.191 &   27.99 &   0.849 \\
        \bottomrule
    \end{tabular}
    }}
    \includegraphics[width=.9\linewidth]{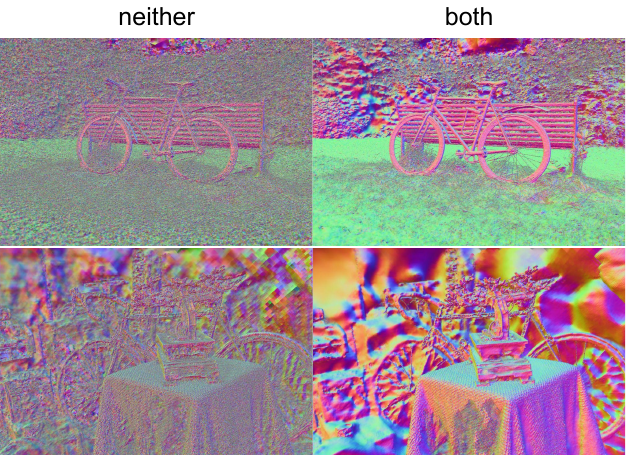}
    \vspace{-.5em}
    \caption{
        {\bf Mesh regularization losses for novel-view synthesis.}
        We also try the normal-depth self-consistency losses for novel-view synthesis task.
        Despite of loss a little in PSNR, the regularization can make the rendered normals much smoother as shown in the visualization.
    }
    \label{tab:supp_nd_loss}
\end{table}

%% file: supp_material/tab/abla_mesh.tex
\begin{table}[h]
    \centering
    {\resizebox{1\linewidth}{!}{
    \begin{tabular}{@{}ccc cc c cc@{}}
        \toprule
        &&& \multicolumn{2}{c}{TnT dataset} && \multicolumn{2}{c}{DTU dataset} \\
        \cmidrule{4-5}\cmidrule{7-8}
        $\loss_{\mathrm{n\text{-}dmed}}$ & $\loss_{\mathrm{n\text{-}dmean}}$ & $K$ & F-score$\uparrow$ & Tr. time$\downarrow$ && Cf.$\downarrow$ & Tr. time$\downarrow$ \\
        \midrule
                   &            & 3 & 0.56 & {\bf 10.1m} && 0.94 & {\bf 5.5m}\\
        \checkmark &            & 3 & 0.59 & {\bf 10.1m} && 0.68 & {\bf 5.5m}\\
                   & \checkmark & 3 & 0.61 & 10.6m && 0.68 & 5.7m\\
        \checkmark & \checkmark & 1 & 0.61 & 10.7m && 0.66 & 5.8m\\
        \checkmark & \checkmark & 2 & 0.61 & 10.8m && {\bf 0.65} & 5.9m\\
        \checkmark & \checkmark & 3 & {\bf 0.62} & 10.9m && {\bf 0.65} & 6.0m\\
        \bottomrule
    \end{tabular}
    }}
    \vspace{-.5em}
    \caption{
        {\bf Mesh regularization losses.}
        We show the results of the mesh regularization losses and the number of sample points when rendering a voxel on a subset of Tanks\&Temples~\cite{tanksandtemples} and DTU~\cite{dtu} datasets.
    }
    \label{tab:supp_mesh_reg}
\end{table}

%% file: supp_material/more_results.tex
\section{More Results}
\label{sec:supp_results}

\paragraph{Comparison with popping-resistant 3DGS variants.}
More comparisons with recent 3DGS variants are in \cref{tab:popping}.
3DGS~\cite{3dgs} has popping artifacts due to the ordering and overlapping issues.
StopThePop~\cite{stopthepop} uses running sort and 3DGRT~\cite{3dgrt} uses ray tracing for accurate ordering but drops FPS by 28\% and 67\% respectively.
EVER~\cite{ever} further handles the Gaussian overlapping cases but with even less FPS.
Our method ensures correct ordering (\cref{sssec:rasterization}) with FPS and quality comparable to the original 3DGS.

\input{tab/popping}

\paragraph{Results on Scannet++.}
Scannet++~\cite{scannetpp} is a large-scale dataset covering various types of indoor scenes.
To reconstruct the bounded indoor environments, we heuristically set the scene center as the camera centroid and the scene radius as twice the maximum camera distance from the centroid. The voxel grid starts at $64^3$ without a background region. Additionally, we implement ray density ascending regularization and a spherical harmonic reset trick, which we find improves results on the public validation set.
The result on the held-out test-set is shown in \cref{tab:scannetpp}.
Our method achieves good results on all metrics.
Some indoor fly-through videos are provided in the released code.

\input{supp_material/tab/scannetpp}

\paragraph{Results breakdown for novel-view synthesis.}
In \cref{tab:supp_quantitative_breakdown}, we show details per-scene comparison with 3DGS~\cite{3dgs} using our base setup.
Our method uses much more primitives (\ie, voxels or Gaussians) compared to 3DGS on all the scenes.
However, our average rendering FPS is still comparable to 3DGS.
We find the FPS is scene-dependent, where we achieve much faster FPS on some of the scenes while slower on the others.
Our method generally uses short training time.
Regarding the quality metrics, our results are typically $-0.2$db PSNR and $-0.01$ SSIM behind 3DGS, while our LPIPS is better on average.

As discussed in the main paper, not only the scene representation itself affects the results, but the optimization and adaptive procedure are also an important factor.
The strategy of adding more Gaussians progressively is not applicable to ours.
We also have not explored to use of the coarse geometry estimated from SfM, while 3DGS uses SfM sparse points for initialization.
As the first attempt of marrying rasterizer with fully explicit sparse voxels for scene reconstruction, there is still future potential for improvement from different aspects.

\input{supp_material/tab/breakdown_nvs}

\paragraph{Results breakdown for mesh reconstruction.}
The F-score and chamfer distance of each scene from the Tanks\&Temples and DTU~\cite{dtu} datasets are provided in \cref{tab:supp_quantitative_breakdown_mesh}.
We only list the two representative NeRF-based methods and two GS-based methods in the result breakdown comparison.
More methods with the average scores are in the main paper.

\input{supp_material/tab/breakdown_mesh}

\paragraph{Synthetic dataset.}
The results on the Synthetic-NeRF~\cite{nerf} dataset is provided in the last section of \cref{tab:supp_quantitative_breakdown_synthetic}.
We achieve good quality, high FPS, and fast training on this dataset.
However, our quality is slightly worse than 3DGS~\cite{3dgs} with slower FPS.
Our development mainly focuses on real-world datasets.
Future work may need more exploration to continue development on this dataset.

\paragraph{More qualitative results}
We show qualitative comparison with 3DGS~\cite{3dgs} on indoor and outdoor scenes in \cref{fig:qual_more_nvs_indoor} and \cref{fig:qual_more_nvs_outdoor}, respectively.
Our visual quality is on par with 3DGS.
We provide the visualization of the raw reconstructed meshes in \cref{fig:qual_more_mesh}.
For quantitative evaluation, we follow previous works to apply mesh cleaning with the provided bounding box or masks.
Despite the good quantitative results for meshes, some apparent artifacts can be observed from the visualization.
In particular, our method focuses more on the geometric details and sometimes over-explains the texture on a flat surface with complex geometry.
Future work may want to model a signed distance field instead of our current density field and introduce surface smoothness regularizers.

\input{supp_material/fig/qual_more_nvs_indoor}
\input{supp_material/fig/qual_more_nvs_outdoor}
\input{supp_material/fig/qual_more_mesh}

%% file: tab/popping.tex
\begin{table}
    \centering
    {\resizebox{1\linewidth}{!}{
    \begin{tabular}{@{}l @{\hskip 6pt}c@{\hskip 6pt}c@{\hskip 6pt}c@{\hskip 6pt}c@{\hskip 6pt} c @{\hskip 6pt}c@{\hskip 6pt}c@{}}
        \toprule
        & \multicolumn{4}{c}{3DGS variants} && \multicolumn{2}{c}{\bf Ours}\\
        \cmidrule{2-5}\cmidrule{7-8}
        Method & 3DGS~\cite{3dgs}\textsuperscript{\textdagger} & StopThePop~\cite{stopthepop}\textsuperscript{\textdagger} & 3DGRT~\cite{3dgrt} & EVER~\cite{ever} && fast-rend & base\\
        {\footnotesize No popping} & & $\triangle$ & $\triangle$ & \checkmark && \checkmark & \checkmark\\
        \midrule
        FPS$\uparrow$ & 
        \silver{131} & 94 & 43\textsuperscript{\textdaggerdbl} & 20\textsuperscript{\textdaggerdbl} &&
        \gold{258} & \brown{121}\\
        LPIPS$\downarrow$ &
        0.257 & 0.251 & \brown{0.248} & \silver{0.233} &&
        0.249 & \gold{0.219}\\
        PSNR$\uparrow$ &
        \silver{27.45} & \brown{27.35} & 27.20 & \gold{27.51} &&
        26.87 & 27.33\\
        SSIM$\uparrow$ & 
        0.815 & 0.816 & \brown{0.818} & \gold{0.825} &&
        0.804 & \silver{0.822}\\
        \bottomrule
        \multicolumn{7}{@{}l@{}}{\footnotesize {{\textdagger} Re-evaluated on our machine using the public code.}}\\
        \multicolumn{7}{@{}l@{}}{\footnotesize {{\textdaggerdbl} We scale the FPS to align their reported 3DGS FPS to our reproduced 3DGS.}}
    \end{tabular}
    }}
    \vspace{-1em}
    \caption{
        {\bf Comparison with 3DGS variants tackling popping artifact on Mip-NeRF360 dataset~\cite{mip-nerf360}.}
        The LPIPS values here are evaluated with the correct intensity scale between $[-1,1]$ following EVER~\cite{ever}.
        3DGRT and EVER use ray tracing approach instead of rasterization.
        EVER solves the Gaussians ordering and overlapping issues but sacrificing more FPS.
    }
    \label{tab:popping}
    \vspace{-.5em}
\end{table}

%% file: supp_material/tab/scannetpp.tex
\begin{table}
    \centering
    {\resizebox{.6\linewidth}{!}{
    \begin{tabular}{@{}lccc@{}}
        \toprule
        Method & LPIPS$\downarrow$ & PSNR$\uparrow$ & SSIM$\uparrow$ \\
        \midrule
        \midrule
        \multicolumn{4}{@{}l}{Small set (12 scenes)} \\
        \midrule
        Plenoxels~\cite{plenoctree} & 0.399 & 22.177 & 0.841 \\
        TensoRF~\cite{tensorf} & 0.404 & 23.524 & 0.857 \\
        INGP~\cite{instant-ngp} & 0.363 & 23.695 & 0.871 \\
        Zip-NeRF~\cite{zipnerf} & 0.320 & \gold{24.630} & \gold{0.887} \\
        Nerfacto~\cite{nerfstudio} & 0.340 & 23.498 & 0.868 \\
        3DGS~\cite{3dgs} & 0.312 & 23.389 & 0.876 \\
        FeatSplat~\cite{featsplat} & \brown{0.303} & \brown{24.177} & 0.880 \\
        RPBG~\cite{rpbg} & \gold{0.271} & 24.005 & \brown{0.882} \\
        Ours & \silver{0.300} & \silver{24.365} & \silver{0.886} \\
        \midrule
        \midrule
        \multicolumn{4}{@{}l}{Full set (50 scenes)} \\
        \midrule
        TensoRF~\cite{tensorf} & 0.406 & 24.022 & 0.850 \\
        Zip-NeRF~\cite{zipnerf} & \brown{0.325} & \gold{25.041} & \gold{0.880} \\
        3DGS~\cite{3dgs} & \silver{0.319} & \brown{23.893} & \brown{0.871} \\
        Ours & \gold{0.313} & \silver{24.709} & \silver{0.874} \\
        \bottomrule
    \end{tabular}
    }}
    \vspace{-.5em}
    \caption{
        {\bf Scannet++~\cite{scannetpp} indoor dataset.}
        The test-set images are not released to prevent overfitting.
        We submit our rendering results to the scannet++ official website for a 3rd-party evaluation.
        The online benchmark is:
        {\footnotesize \url{https://kaldir.vc.in.tum.de/scannetpp/benchmark/nvs}}.
        In average, our training time is 12 minutes per scene; our rendering FPS is 197 at $1752 \times 1168$ resolutions.
        Voxel size statistic is:
        13.61\% \textless3mm, 19.25\% 3-5mm, 32.43\% 5mm-1cm, 23.31\% 1-2cm, 6.66\% 2-3cm, 4.73\% \textgreater3cm.
        We do not use the sparse points prior from COLMAP~\cite{colmap} in this submission.
    }
    \label{tab:scannetpp}
\end{table}

%% file: supp_material/tab/breakdown_nvs.tex
\begin{table*}[t]
    \centering
    {\resizebox{1\linewidth}{!}{
    \begin{tabular}{@{}l cc c cc c cc c cc c cc c cc@{}}
        \toprule
        &
        \multicolumn{2}{c}{FPS$\uparrow$} &&
        \multicolumn{2}{c}{Tr. time$\downarrow$ (mins)} && 
        \multicolumn{2}{c}{LPIPS$\downarrow$} && 
        \multicolumn{2}{c}{PSNR$\uparrow$} && 
        \multicolumn{2}{c}{SSIM$\uparrow$} && 
        \multicolumn{2}{c}{\# prim.$\downarrow$} \\
        \cmidrule{2-3}\cmidrule{5-6}\cmidrule{8-9}\cmidrule{11-12}\cmidrule{14-15}\cmidrule{17-18}
        Scene &
        3DGS & ours &&
        3DGS & ours &&
        3DGS & ours &&
        3DGS & ours &&
        3DGS & ours &&
        3DGS & ours \\
        \midrule
        \midrule
        \multicolumn{18}{@{}l}{MipNeRF-360~\cite{mip-nerf360} indoor scenes} \\
        \midrule
bonsai     & {\bf 215} & 128      && 18.3 & {\bf 15.3}    && 0.204 & {\bf 0.171}  && {\bf 31.89} & 31.51  &&      0.942  & {\bf 0.944}  && {\bf 1.2M} & 6.6M    \\
counter    & {\bf 160} &  85      && 20.7 & {\bf 18.7}    && 0.199 & {\bf 0.176}  && {\bf 29.03} & 28.72  && {\bf 0.909} &      0.905   && {\bf 1.2M} & 8.4M    \\
kitchen    & {\bf 128} &  78      && 25.0 & {\bf 17.8}    && 0.126 & {\bf 0.112}  && {\bf 31.47} & 31.29  &&      0.927  & {\bf 0.934}  && {\bf 1.8M} & 9.2M    \\
room       & {\bf 153} & 131      && 21.1 & {\bf 17.1}    && 0.218 & {\bf 0.185}  && {\bf 31.44} & 31.10  &&      0.919  & {\bf 0.924}  && {\bf 1.5M} & 8.9M    \\
        \midrule
        \midrule
        \multicolumn{18}{@{}l}{MipNeRF-360~\cite{mip-nerf360} outdoor scenes} \\
        \midrule
bicycle    & 72 & {\bf 147}       && 31.9 & {\bf 13.5}    && 0.211 & {\bf 0.190}  && 25.18 & {\bf 25.29}  && 0.765 & {\bf 0.773}  && {\bf 6.1M} & 9.2M    \\
garden     & 81 & {\bf 118}       && 33.1 & {\bf 12.4}    && 0.107 & {\bf 0.106}  && {\bf 27.39} & 27.31  && {\bf 0.867} & 0.865  && {\bf 5.9M} & 9.6M    \\
stump      & 110 & {\bf 129}      && 25.5 & {\bf 13.0}    && 0.216 & {\bf 0.206}  && {\bf 26.61} & 26.38  && {\bf 0.772} & 0.769  && {\bf 4.9M} & 9.2M    \\
treehill   & 123 & {\bf 157}      && 22.4 & {\bf 13.7}    && 0.327 & {\bf 0.262}  && 22.47 & {\bf 22.74}  && 0.632 & {\bf 0.646}  && {\bf 3.7M} & 9.4M    \\
flowers    & {\bf 137} & 120      && 22.0 & {\bf 14.4}    && 0.335 & {\bf 0.268}  && 21.57 & {\bf 21.72}  && 0.606 & {\bf 0.637}  && {\bf 3.6M} & 9.4M    \\
        \midrule
        \midrule
        \multicolumn{18}{@{}l}{DeepBlending~\cite{deepblending} indoor scenes} \\
        \midrule
drjohnson  & 116 & {\bf 297}      && 25.0 & {\bf 8.7}     && 0.244 & {\bf 0.242}  && 29.11 & {\bf 29.22}  && {\bf 0.901} & 0.892  && {\bf 3.3M} & 6.8M    \\
playroom   & 163 & {\bf 308}      && 19.7 & {\bf 7.4}     && 0.244 & {\bf 0.211}  && 30.08 & {\bf 30.53}  && {\bf 0.907} & 0.900  && {\bf 2.3M} & 6.3M    \\
        \midrule
        \midrule
        \multicolumn{18}{@{}l}{Tanks\&Temples~\cite{tanksandtemples} outdoor scenes} \\
        \midrule
train      & {\bf 206} & 127      && {\bf 11.3} & 11.4    && 0.206 & {\bf 0.186}  && {\bf 22.11} & 21.26  && {\bf 0.816} & 0.813  && {\bf 1.1M} & 8.3M    \\
truck      & {\bf 154} & 129      && 16.3 & {\bf 11.0}    && 0.147 & {\bf 0.100}  && {\bf 25.40} & 25.00  && 0.882 & {\bf 0.888}  && {\bf 2.6M} & 9.0M    \\
        \bottomrule
    \end{tabular}
    }}
    \vspace{-.5em}
    \caption{
        {\bf Real-world datasets for per-scene side-by-side comparison with 3DGS~\cite{3dgs}.}
        Our result here is the base setup.
        We show average results of our 2x faster rendering and 3x faster training variants in the main paper.
    }
    \label{tab:supp_quantitative_breakdown}
\end{table*}

\begin{table*}[t]
    \centering
    {\resizebox{1\linewidth}{!}{
    \begin{tabular}{@{}l cc c cc c cc c cc c cc c cc@{}}
        \toprule
        &
        \multicolumn{2}{c}{FPS$\uparrow$} &&
        \multicolumn{2}{c}{Tr. time$\downarrow$ (mins)} && 
        \multicolumn{2}{c}{LPIPS$\downarrow$} && 
        \multicolumn{2}{c}{PSNR$\uparrow$} && 
        \multicolumn{2}{c}{SSIM$\uparrow$} && 
        \multicolumn{2}{c}{\# prim.$\downarrow$} \\
        \cmidrule{2-3}\cmidrule{5-6}\cmidrule{8-9}\cmidrule{11-12}\cmidrule{14-15}\cmidrule{17-18}
        Scene &
        3DGS & ours &&
        3DGS & ours &&
        3DGS & ours &&
        3DGS & ours &&
        3DGS & ours &&
        3DGS & ours \\
        \midrule
        \midrule
        \multicolumn{18}{@{}l}{Synthetic-NeRF~\cite{nerf} object scenes} \\
        \midrule
chair      & {\bf 418} & 197      && 5.4 & {\bf 4.6}      && {\bf 0.012} & 0.013  && 35.89 & {\bf 35.91}  && {\bf 0.987} & 0.986  && {\bf 0.3M} & 3.0M    \\
drums      & {\bf 406} & 241      && 5.8 & {\bf 4.1}      && {\bf 0.037} & 0.043  && {\bf 26.16} & 26.09  && {\bf 0.955} & 0.947  && {\bf 0.3M} & 2.3M    \\
ficus      & {\bf 476} & 360      && 5.0 & {\bf 3.1}      && {\bf 0.012} & 0.014  && {\bf 34.85} & 34.37  && {\bf 0.987} & 0.984  && {\bf 0.3M} & 1.3M    \\
hotdog     & {\bf 596} & 218      && 5.2 & {\bf 4.8}      && 0.020 & {\bf 0.019}  && {\bf 37.67} & 37.42  && {\bf 0.985} & 0.984  && {\bf 0.1M} & 2.8M    \\
lego       & {\bf 415} & 156      && 5.9 & {\bf 5.8}      && {\bf 0.015} & 0.016  && {\bf 35.77} & 35.54  && {\bf 0.983} & 0.981  && {\bf 0.3M} & 4.3M    \\
materials  & {\bf 575} & 213      && 5.3 & {\bf 4.5}      && {\bf 0.034} & 0.037  && {\bf 30.01} & 30.00  && {\bf 0.960} & 0.954  && {\bf 0.3M} & 2.7M    \\
mic        & {\bf 344} & 328      && 5.6 & {\bf 3.0}      && {\bf 0.006} & 0.007  && 35.38 & {\bf 36.00}  && 0.991 & {\bf 0.992}  && {\bf 0.3M} & 1.1M    \\
ship       & {\bf 254} & 111      && {\bf 7.9} & 8.6      && 0.107 & {\bf 0.106}  && {\bf 30.92} & 30.38  && {\bf 0.907} & 0.886  && {\bf 0.3M} & 5.7M    \\
        \bottomrule
    \end{tabular}
    }}
    \vspace{-.5em}
    \caption{
        {\bf Synthetic object-centric dataset for per-scene side-by-side comparison with 3DGS~\cite{3dgs}.}
        Our overall quality is slightly worse than 3DGS on this dataset while the synthetic object-centric scenario is off our main focus.
    }
    \label{tab:supp_quantitative_breakdown_synthetic}
\end{table*}

%% file: supp_material/tab/breakdown_mesh.tex
\begin{table*}[t]
    \centering
    {\resizebox{1\linewidth}{!}{
    \begin{tabular}{@{}l cccccc c ccccccccccccccc@{}}
        \toprule
        & \multicolumn{6}{c}{Tanks\&Temples F-score$\uparrow$} && \multicolumn{15}{c}{DTU Chamfer Cistance$\downarrow$} \\
        \cmidrule{2-7}\cmidrule{9-23}
        Method &
        Barn & Caterpillar & Courthouse & Ignatius & Meetingroom & Truck &&
        24 & 37 & 40 & 55 & 63 & 65 & 69 & 83 & 97 & 105 & 106 & 110 & 114 & 118 & 122\\
        \midrule
        NeuS         &
        0.29 & 0.29 & 0.17 & \silver{0.83} & \silver{0.24} & 0.45 &&
        1.00 & 1.37 & 0.93 & 0.43 & 1.10 & \silver{0.65} & \silver{0.57} & 1.48 & \silver{1.09} & 0.83 & \silver{0.52} & 1.20 & \silver{0.35} & \silver{0.49} & 0.54\\
        Neuralangelo &
        \gold{0.70} & \gold{0.36} & \silver{0.28} & \gold{0.89} & \gold{0.32} & \silver{0.48} &&
        \gold{0.37} & \gold{0.72} & \gold{0.35} & \gold{0.35} & \gold{0.87} & \gold{0.54} & \gold{0.53} & \gold{1.29} & \gold{0.97} & \silver{0.73} & \gold{0.47} & \gold{0.74} & \gold{0.32} & \gold{0.41} & \gold{0.43}\\
        3DGS         &
        0.13 & 0.08 & 0.09 & 0.04 & 0.01 & 0.19 &&
        2.14 & 1.53 & 2.08 & 1.68 & 3.49 & 2.21 & 1.43 & 2.07 & 2.22 & 1.75 & 1.79 & 2.55 & 1.53 & 1.52 & 1.50\\
        2DGS         &
        \silver{0.41} & 0.23 & 0.16 & 0.51 & 0.17 & 0.45 &&
        \silver{0.48} & 0.91 & \silver{0.39} & 0.39 & 1.01 & 0.83 & 0.81 & 1.36 & 1.27 & 0.76 & 0.70 & 1.40 & 0.40 & 0.76 & 0.52\\
        ours         &
        0.35 & \silver{0.33} & \gold{0.29} & 0.69 & 0.19 & \gold{0.54} &&
        0.61 & \silver{0.74} & 0.41 & \silver{0.36} & \silver{0.93} & 0.75 & 0.94 & \silver{1.33} & 1.40 & \gold{0.61} & 0.63 & \silver{1.19} & 0.43 & 0.57 & \silver{0.44}\\
        \bottomrule
    \end{tabular}
    }}
    \vspace{-.5em}
    \caption{
        {\bf Result breakdown on Tanks\&Temples~\cite{tanksandtemples} and DTU~\cite{dtu} datasets.}
    }
    \label{tab:supp_quantitative_breakdown_mesh}
\end{table*}

%% file: supp_material/fig/qual_more_nvs_indoor.tex
\begin{figure*}[h]
    \centering
    \includegraphics[width=.85\linewidth]{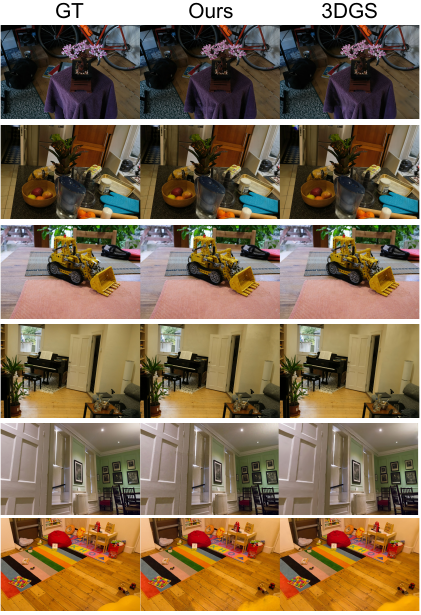}
    \caption{
    {\bf Qualitative novel-view rendering results on-par with 3DGS.}
    }
    \label{fig:qual_more_nvs_indoor}
\end{figure*}

%% file: supp_material/fig/qual_more_nvs_outdoor.tex
\begin{figure*}[h]
    \centering
    \includegraphics[width=.85\linewidth]{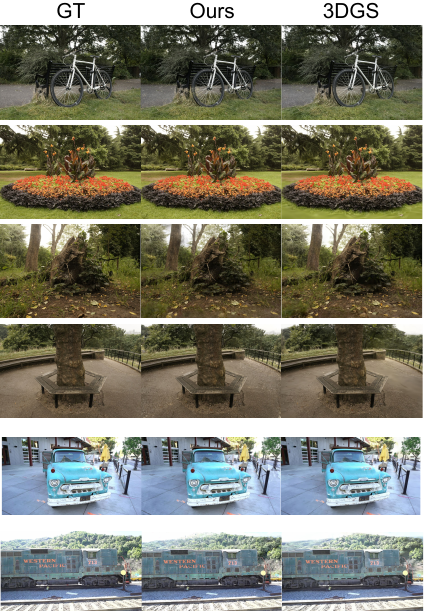}
    \caption{
    {\bf Qualitative novel-view rendering results on-par with 3DGS.}
    }
    \label{fig:qual_more_nvs_outdoor}
\end{figure*}

%% file: supp_material/fig/qual_more_mesh.tex
\begin{figure*}[h]
    \centering
    \includegraphics[width=\linewidth]{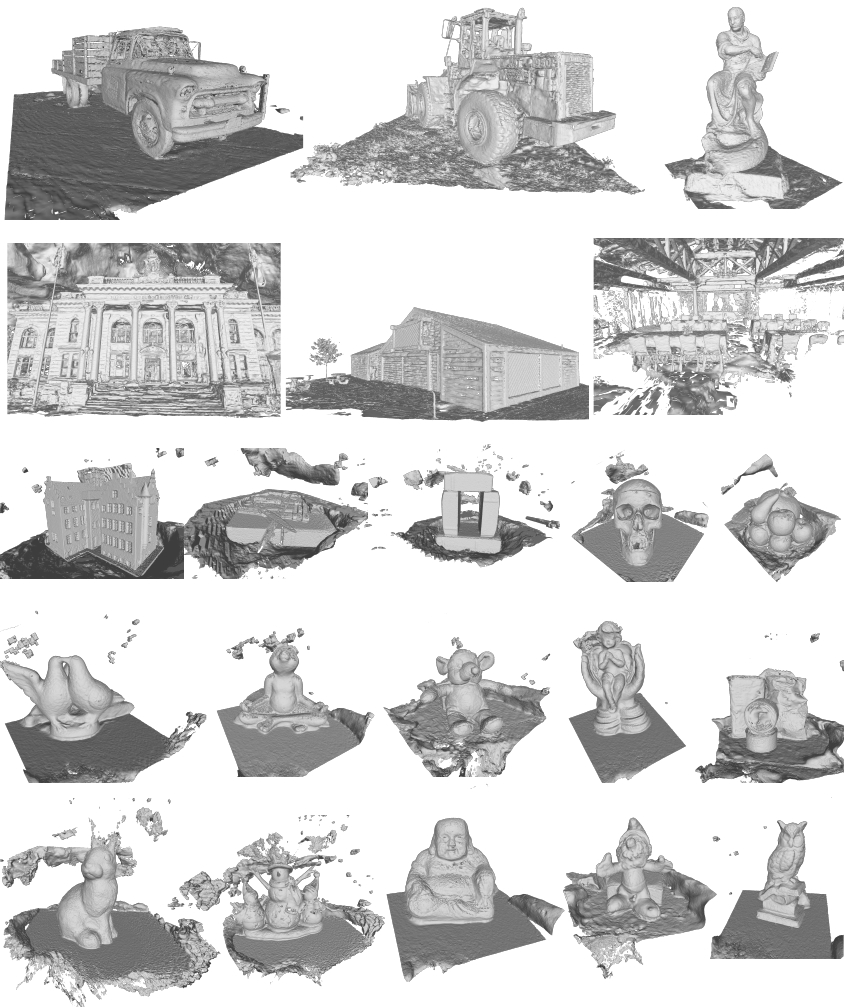}
    \caption{
    {\bf Qualitative results of the reconstructed mesh.}
    }
    \label{fig:qual_more_mesh}
\end{figure*}